%% file: Main.tex
\documentclass{article}
\usepackage{appendix}
\usepackage{float}
\usepackage[T1]{fontenc}
\usepackage{lmodern}
\usepackage{tikz}
\usetikzlibrary{patterns}
\usetikzlibrary{shapes.symbols, shapes.callouts}
 \usetikzlibrary{shapes.geometric, arrows, positioning}
\usetikzlibrary{positioning}

\definecolor{tokenfill}{HTML}{E1F5EE}
\definecolor{tokenstroke}{HTML}{0F6E56}
\definecolor{tokentext}{HTML}{04342C}
\definecolor{coral}{HTML}{D85A30}
\definecolor{coralfill}{HTML}{FAECE7}
\definecolor{coraltext}{HTML}{993C1D}
\definecolor{cellfill}{HTML}{F1EFE8}
\definecolor{cellstroke}{HTML}{888780}

\usepackage[T1]{fontenc}
\usepackage{lmodern}
\usepackage[margin=1in]{geometry}
 
\usepackage{graphicx}                    
\usepackage{booktabs}                    
\usepackage[flushleft]{threeparttable}   
\usepackage{cancel} 
\usepackage{amsmath} 
 
\usepackage{graphicx}

\usepackage{algorithm}
\usepackage{algorithmic}
\usepackage[english]{babel}
\usepackage{xcolor}

\usepackage{amssymb}
\usepackage{standalone}
\usepackage{siunitx}
\PassOptionsToPackage{hyphens}{url}\usepackage{hyperref}
\usepackage{cleveref}
\usepackage[utf8]{inputenc}
\usepackage[right]{lineno}
\usepackage{csquotes}
\usepackage{booktabs}
\usepackage{longtable}
\usepackage{adjustbox}
\usepackage{array}
\usepackage{url}
\usepackage{titlesec}
\usepackage{authblk}
\usepackage{xcolor} 

\titleformat{\subsection}
  {\mdseries\itshape\large} 
  {\thesubsection}{1em}{} 

\usepackage[english]{babel}
\usepackage[style=authoryear,backend=biber,natbib=true,maxcitenames=2,uniquelist=false]{biblatex}
\makeatletter
\NewBibliographyString{availableat}
\makeatother

\DefineBibliographyStrings{english}{
  availableat = {available at},
}
\DeclareNameAlias{sortname}{family-given}
\DeclareNameAlias{default}{family-given}

\renewbibmacro{in:}{}
\DeclareFieldFormat[article]{title}{\mkbibquote{#1}\addcomma}
\DeclareFieldFormat[book]{title}{\mkbibemph{#1}\addcomma}
\DeclareFieldFormat[bookinbook]{title}{\mkbibemph{#1}\addcomma}
\DeclareFieldFormat[inbook]{title}{\mkbibquote{#1}\addcomma}
\DeclareFieldFormat[incollection]{title}{\mkbibquote{#1}\addcomma}
\DeclareFieldFormat[inproceedings]{title}{\mkbibquote{#1}\addcomma}
\DeclareFieldFormat[manual]{title}{\mkbibemph{#1}\addcomma}
\DeclareFieldFormat[misc]{title}{\mkbibemph{#1}\addcomma}
\DeclareFieldFormat[thesis]{title}{\mkbibemph{#1}\addcomma}
\DeclareFieldFormat[unpublished]{title}{\mkbibquote{#1}\addcomma}
\DeclareFieldFormat[patent]{title}{\mkbibemph{#1}\addcomma}
\DeclareFieldFormat[report]{title}{\mkbibemph{#1}\addcomma}
\DeclareFieldFormat[online]{title}{\mkbibquote{#1}\addcomma}
\DeclareFieldFormat[software]{title}{\mkbibemph{#1}\addcomma}
\DeclareFieldFormat[booklet]{title}{\mkbibemph{#1}\addcomma}
\DeclareFieldFormat[periodical]{title}{\mkbibemph{#1}\addcomma}
\DeclareFieldFormat[standard]{title}{\mkbibemph{#1}\addcomma}

\DeclareFieldFormat[article]{journaltitle}{\iffieldundef{shortjournal}{\mkbibemph{#1}\addcomma}{\mkbibemph{\printfield{shortjournal}}\addcomma}}
\DeclareFieldFormat{volume}{\bibstring{volume}~#1}
\DeclareFieldFormat{number}{\bibstring{number}~#1}

\DefineBibliographyStrings{english}{
  volume = {Vol.},
  number = {No.}
}

\renewbibmacro*{volume+number+eid}{%
  \printfield{volume}%
  \setunit*{\addspace}%
  \printfield{number}%
  \setunit{\addcomma\space}%
  \printfield{eid}}

\renewbibmacro*{journal+issuetitle}{%
  \usebibmacro{journal}%
  \setunit*{\addcomma\space}%
  \usebibmacro{volume+number+eid}%
  \setunit{\addcomma\space}%
  \usebibmacro{issue+date}}

\renewbibmacro*{publisher+location+date}{%
  \printlist{publisher}%
  \iflistundef{location}
    {\setunit*{\addcomma\space}}
    {\setunit*{\addcolon\space}}%
  \printlist{location}%
  \setunit*{\addcomma\space}%
  \usebibmacro{date}}

\DeclareCiteCommand{\cite}[\mkbibparens]
  {\usebibmacro{prenote}}
  {\usebibmacro{citeindex}%
   \usebibmacro{cite}}
  {\multicitedelim}
  {\usebibmacro{postnote}}

\renewbibmacro*{cite:labelyear+extrayear}{%
  \iffieldundef{labelyear}
    {}
    {\printtext[bibhyperref]{%
       \printfield{labelyear}%
       \printfield{extrayear}}}}

\renewbibmacro*{cite:labeldate+extradate}{%
  \iffieldundef{labelyear}
    {}
    {\printtext[bibhyperref]{%
       \printfield{labelyear}%
       \printfield{extradate}}}}

\AtEveryBibitem{
  \clearfield{month}
  \clearfield{day}
  \ifentrytype{book}{
    \clearlist{location}
  }{}
}

\DefineBibliographyStrings{english}{
  andothers = {\textit{et al.},}
}

\DeclareFieldFormat[article]{volume}{\bibstring{jourvol}\addnbspace #1}
\DeclareFieldFormat[article]{number}{\bibstring{number}\addnbspace #1}
\DeclareFieldFormat[article]{volume}{Vol. #1}
\DeclareFieldFormat[article]{number}{No. #1}

\DeclareFieldFormat{url}{\bibstring{availableat}\addcolon\space\url{#1}}
\DeclareFieldFormat{urldate}{\mkbibparens{accessed \addspace#1}}

\DeclareFieldFormat{urldate}{%
  \mkbibparens{accessed\space%
    \thefield{urlday}\addspace%
    \mkbibmonth{\thefield{urlmonth}}\addspace%
    \thefield{urlyear}}}

\crefformat{figure}{#2Figure~#1#3}
\Crefformat{figure}{#2Figure~#1#3}
\crefformat{table}{#2Table~#1#3}
\Crefformat{table}{#2Table~#1#3}
\crefformat{section}{#2Section~#1#3}
\Crefformat{section}{#2Section~#1#3}

\author[1]{Shibingfeng Zhang}
\author[1]{Edoardo Caraffa}
\author[1]{Annafelicia Zuffrano}
\author[1]{Maddalena Modesti}
\author[1,2]{Giovanni Colavizza}
 \affil[1]{Department of Classical Philology and Italian Studies, University of Bologna}
\affil[2]{Centre for Digital and Computational Humanities, University of Copenhagen}
\date{}
\title{Text Restoration of Ancient Documents with Language Models}

\begin{document}
\maketitle

\begin{abstract}
\textbf{Purpose} - This study investigates the feasibility of restoring missing text caused by physical lacunae in damaged ancient manuscripts using language models.

\textbf{Methodology} - The study proposes different scenarios to replicate real-world conditions. Language models of different architectures are applied according to their suitability to each scenario. We also propose several decoding strategies that further enhance performance and address the discrepancy between lacuna boundaries and the models' tokenization schemes.

\textbf{Findings} - The results reveal that text restoration of these documents cannot be fully automated, but it can serve as a useful tool to assist paleographers in their work. Model performance varies greatly depending on which structural part of the document needs to be restored and whether the character length of missing text is available.

\textbf{Originality} - This is the first study and to analyze model performance on formulaic and non-formulaic content and the impact of lacuna length awareness in manuscript restoration. Both are recurring challenges in paleographers' manual restoration work. Through systematic comparison and both qualitative and quantitative analysis of different models' performance under varying settings, this study offers a guideline for developing assistive tools to support paleographers.
\\
\\

\textbf{Keywords:} Text restoration; Latin manuscripts; Palaeography; Diplomatics; Large Language Models; Natural Language Processing; Digital Humanities
\end{abstract}

\section{Introduction}
\label{sec:introduction}

Ancient written texts constitute a fundamental part of our cultural heritage, providing essential evidence for the reconstruction of past societies, institutions, and intellectual traditions. Preserved across a wide range of documentary and textual traditions, including manuscripts, archival records, inscriptions, and epigraphic sources, these materials are often incomplete or damaged due to the long-time conservation. Missing text is therefore a pervasive problem in the study of historical written sources, arising from physical deterioration, lacunae, illegible passages, or the loss of portions of the original document. The reconstruction of such missing passages is consequently a central challenge for scholars working with historical texts, requiring them to combine linguistic, textual, material, and contextual evidence.
Recent advances in computational methods and artificial intelligence have opened new possibilities for addressing this challenge~\cite{assael-etal-2019-restoring,lazar-etal-2021-filling,shen-etal-2020-blank}, with many applications focusing on text restoration of epigraphy. Automated approaches to text restoration can potentially assist scholars in reconstructing lacunae across large and heterogeneous corpora, overcoming some of the limitations imposed by the traditionally time-consuming and highly specialized nature of this task. Yet the variety of historical languages, textual genres, documentary forms, and writing practices makes it difficult to develop universally applicable restoration methods~\cite{sommerschield-etal-2023-machine}. Effective approaches must therefore account not only for linguistic and textual regularities, but also for the specific characteristics of the sources to which they are applied. This calls for robust restoration pipelines that keep the scholars in the loop, using automated methods to propose and rank candidate reconstructions that domain experts can then evaluate, refine, or reject. This positions computational text restoration not as a standalone technology, but as a tool developed in close collaboration with the disciplines, such as diplomatics and palaeography, that have long specialized in exactly this kind of reconstructive work.

In this respect, the rapid development of natural language processing in recent years is especially relevant. Several studies have attempted to create language models capable of handling text in historical languages such as Latin~\cite{bamman2020latin}, ancient Greek~\cite{riemenschneider2023exploring}, Old Italian~\cite{aprosio2022bertoldo}, and ancient Chinese~\cite{wang2023gujibert}. This growing research landscape allows us to investigate the feasibility of employing language models to reconstruct textual gaps in the context of historical manuscript, to facilitate the work of paleographers. Such methods aim to enhance its effectiveness by helping to formulate restoration hypotheses for incomplete sections.

For this purpose, we use a dataset consisting of around 1,200 notarial documents produced during the Late Middle Ages in Bologna~\cite{Modesti2012studi}. As these documents have never been publicly available, they are guaranteed to be unseen by any existing language model, ensuring a fair comparison. The Bolognese notarial tradition represents a privileged point of observation for analyzing the development of documentary practices and the relationship between documentary form, legal function, and social context \cite{Fasoli1977Notaio, Tamba1998instrumenti1}. Their structure also enables a systematic comparison between restoration performance on formulaic versus non-formulaic content, which is a distinction applicable to many other ancient text traditions, making this work highly transferable. For instance, the same methodological distinction is used in fields such as epigraphy, where densely packed, highly standardized abbreviations coexist with unique context-specific nouns and functions \cite{assael-etal-2019-restoring, Soffiantini2024Latin}. Similarly, within medieval manuscript studies, liturgical texts rely heavily on fixed, highly formulaic structures designed for oral repetition and memorization, contrasting sharply with their unique narrative or marginal annotations \cite{DeshussesDarragon1982}.

Various language models, ranging from smaller-scale encoder-based models to large-scale general-purpose LLMs, are investigated under different settings according to their suitability as determined by model architecture.



The objectives of this study are:
\begin{enumerate}
    \item To investigate the feasibility of using language models to reconstruct missing or illegible content in ancient texts, and to determine the extent to which reliable restoration can be achieved.
    \item To compare and evaluate the performance of different language model architectures and sizes across multiple experimental settings, in order to identify best practices for text restoration of different scenarios.
\end{enumerate}

The contributions of this work are as follows:
\begin{enumerate}

    \item We propose and compare several strategies for reconciling character-level lacuna lengths with the token-level units used by language models, including scenarios where the expected output length is unknown.
    \item Systematic experiments and analysis are conducted across models and text categories, offering insights that can inform future work on text restoration in similarly under-resourced or historically specialized domains.
    \item A set of practical fine-tuned language models is provided to support paleographers and diplomatists in their text restoration work.
\end{enumerate}

\section{Related Work}
\label{sec:related}
This section presents a comprehensive literature review in two parts. The first part reviews text restoration of historical texts, covering both traditional manual restoration performed by professionals and automatic restoration methods based on language technologies, with a special focus on works that employ language models or large language models (LLMs) for the restoration of manuscripts or inscriptions, along with their methodologies and results. The second part examines language models potentially suitable for the restoration of Latin manuscripts, as relevant to the present work.

\subsection{Text Restoration}
Text restoration refers to the process of recovering lost or illegible parts of an ancient text that have become fragmented or damaged due to the deterioration or destruction of writing supports over the time~\cite{sommerschield-etal-2023-machine}. Traditionally, the restoration of historical documents is linked to the methodological contributions of disciplines such as paleography and diplomatics: paleography provides the tools necessary to understand historical writing practices and the material realization of texts, while diplomatics focuses on the form, structure, production, transmission, and documentary function of written records~\cite{Schiaparelli1972Diplomatica, Sickel1975Beitraege, Bresslau1998, Valenti2000Scritti}. Professionals typically rely on two main procedures. The analogical procedure is based on comparison with the scribe's characteristic writing habits and with parallel passages drawn from formularies, earlier documents, normative texts, and other evidence attributable to the same legal or documentary context. The statistical procedure is based on a quantitative assessment of the occurrences of a word or textual sequence within the output of the same scribe, chancery, or epigraphy, used when analogical comparison alone is insufficient~\cite{chiesa2012elementi}.

Recent advances in natural language processing have led to numerous studies applying language technologies to the task of text restoration. A common practice to address this challenge is adapting a general-use pre-trained language model to a specific domain, as available resources for ancient languages tend to be scarce. In general, studies in this field assume that the length of missing text caused by a lacuna is estimated from the size of the physical damage. This information is either used as explicit input to the model or implicitly reflected as the quantity of token masks in the input. In the latter case, the tokenizer of the language model in question is trained in such a way that each character in the language corresponds to one token. For example, ~\citet{lazar-etal-2021-filling} investigated the restoration of Ancient Akkadian inscriptions that were originally written on clay tablets using the BERT model~\cite{devlin-etal-2019-bert}. They fine-tuned multilingual BERT-cased, and the performance of the model surpassed that of the BERT and LSTM~\cite{fetaya2020restoration} models trained from scratch on the same dataset. In this case, text restoration is formulated as a standard masked language modeling task: each cuneiform sign in the clay tablet is one token, and each missing sign is represented as one mask in the input. Blank Language Model~\cite{shen-etal-2020-blank} is an encoder-based model that specializes in text infilling. Researchers tuned and tested the Blank Language Model on the PHI-ML dataset~\cite{assael-etal-2019-restoring}, made of fragments of ancient Greek inscriptions, under the assumption of known character length of the lacuna. The achieved results that are on par with Pythia~\cite{assael-etal-2019-restoring}, a sequence-to-sequence model specialized in ancient Greek inscription restoration.~\citet{wang2025chinese} proposed a multimodal system to address the challenge of text restoration in ancient Chinese inscriptions combining both visual and textual inputs. The textual processing component's backbone is SikuRoBERTa~\cite{wang2022construction}, which is a RoBERTa~\cite{liu2019roberta} based model pre-trained on on the Siku Quanshu (Complete Library of the Four Treasuries), a large-scale corpus of traditional Chinese classical texts. During inference, each missing character on the inscription is represented as one mask. ~\citet{assael2025contextualizing} proposed Aeneas model, which is equipped with a text restoration module and, unlike the above-mentioned methods, is capable of operating both when the length of the missing text is known and when it is unknown, thereby offering maximum flexibility during inference. The model was trained from scratch on the data assembled from several epigraphic data sources~\cite{cowey2019epigraphic,panciera2019edr,hermankova2022edcs} in Latin and ancient Greek and spanning inscriptions from the seventh century BCE to the eighth century CE.

Apart from the aforementioned methods that follow the ``pre-train and fine-tune'' paradigm, another potential approach for text restoration in historical language is the use of large-scale language models, such as DeepSeek~\cite{xu2026deepseek} and LLaMA~\cite{grattafiori2024llama3herdmodels}, leveraging their emerging generalization to unseen tasks. However, to the best of our knowledge, this approach remains unexplored for text restoration in historical texts. Previous studies have demonstrated that providing task-specific examples within prompts can substantially improve LLM performance~\cite{brown2020language}. Accordingly, the present study investigates the effectiveness of few-shot prompting for historical text restoration. Furthermore, drawing inspiration from established procedures in manual restoration, the study provides LLMs with word concordances as an additional source of contextual information. Implementation details are provided in Section~\ref{general llms}.

Building on this current research landscape, the present work investigates both the common known-length mode and the less common unknown-length mode, in order to examine the tradeoff between the effort required from diplomatists to estimate the character length of a lacuna and the resulting restoration accuracy. In particular, given the largely unexplored potential of large-scale language models for text restoration, the feasibility of using LLMs for this purpose is also assessed, with prompts constructed by drawing on established manual restoration practices. The following section examines the language models for Latin that can be used for this task.

\subsection{Language Models for Latin}
\label{lmforlatin}
Given the relatively small size of the notarial documents dataset, it would be ideal to opt for models pre-trained on Latin and tune them to the specific purpose of manuscript restoration. Several resources could be employed for this purpose and are reviewed below.

~\citet{bamman2020latin} proposed LatinBERT, the first contextual language model for Latin. The model follows the BERT~\cite{devlin-etal-2019-bert} architecture and was pre-trained on texts drawn from various sources, mainly the Internet Archive.~\footnote{\url{https://archive.org/}} The model achieved state-of-the-art results on several Latin POS tagging and Latin Universal Parsing tasks. Following this work, \citet{riemenschneider2023exploring} introduce a series of general-purpose mono- and multilingual language models trained from scratch on ancient Greek, Latin, and English, presenting the first generative models for these languages together with corresponding encoder-only models. Among this series, four models are of particular interest to the current work: LaBERTa, PhilBERTa, LaTa, and PhilTa.
LaBERTa and PhilBERTa are encoder-based models, both adopting the same architecture as RoBERTa~\cite{zhuang-etal-2021-robustly} and differing only in their training data. LaBERTa was trained exclusively on Corpus Corporum, a Latin corpus comprising roughly 167 million tokens, while PhilBERTa, the multilingual counterpart, was trained on Corpus Corporum together with 185 million tokens of Ancient Greek and 212 million tokens of antiquity-related English text, such as translations of works originally written in Latin or ancient Greek. LaTa and PhilTa~\cite{riemenschneider2023exploring} follow the same monolingual/multilingual contrast, but are encoder-decoder models based on the T5~\cite{raffel2020exploring} architecture, pre-trained from scratch using the T5-base configuration on an MLM objective. As with the previous pair, the two models differ only in training data, drawn from the same underlying material as LaBERTa and PhilBERTa: PhilTa is trained on Latin alone, while LaTa is trained on Latin, ancient Greek, and English.
In both pairs, this monolingual-versus-multilingual setting was designed to examine the impact of multilingual pre-training data on the performance of language models for historical languages. Across several tasks targeting the semantic, syntactic, and morphological properties of Latin, ~\citet{riemenschneider2023exploring} found that the monolingual and multilingual models performed comparably, with no statistically significant differences on most tasks, and suggested that this could be due to the relatively small size of the test set, leaving the effect of multilingual pretraining an open question. To investigate whether multilinguality aids the text restoration of Latin manuscripts, the present study experiments with both pairs, but assigns them to different settings based on their architectural properties. LaBERTa and PhilBERTa are used for the length-known setting. By formulating text restoration as an MLM task, the length of the output can be controlled. LaTa and PhilTa, by contrast, are used for the length-unknown setting, since T5-style pre-training uses sentinel tokens to predict a variable number of tokens per mask. Further details on implementation of these models are presented in Section~\ref{Length-known Setting} and~\ref{Length-unknown Setting}.

Recently, ~\citet{assael2025contextualizing} proposed Aeneas, a generative neural network specifically developed for the contextualization of Latin inscriptions. It jointly addresses two tasks: (1) \emph{textual restoration}, which takes the incomplete text of an inscription as input and infers the missing characters; (2) \emph{metadata attribution}, which takes the text and image as input to predict the geographical and chronological origin of the inscription. Its architecture processes the transcribed text through a T5 transformer decoder~\cite{raffel2020exploring}, while a vision network encodes the corresponding inscription image for the metadata attribution task. The resulting representations are then passed to a set of task-specific feed-forward neural network heads. The model was trained from scratch on the data assembled from several epigraphic data sources~\cite{cowey2019epigraphic,panciera2019edr,hermankova2022edcs} and spanning inscriptions from the seventh century BCE to the eighth century CE. The dataset comprises 176,861 inscriptions and contains around 16 million characters. For the present work, only the T5 decoder and the feed-forward neural network head dedicated to text restoration are employed. A particularity of this model is that it supports both lacuna length-known and -unknown scenarios for text restoration, making it very suitable for this study. Further details of implementation of this model are presented in Section~\ref{aeneas} and Section~\ref{Aeneasunk}.

\section{Data}
\label{sec:data}


The dataset utilized in this study comprises 1,184 notarial documents produced in Latin at Bologna, Italy, between the early eleventh and late thirteenth centuries, covering a period of approximately 280 years. Since 1937, when Giorgio Cencetti proposed the creation of the \textit{Codice diplomatico bolognese}, a comprehensive edition intended to systematically transcribe and preserve the city’s medieval documentary heritage, research has progressed significantly, with the successful academic publication of 10th- and 11th-century documents \cite{Cencetti1977carteX, Feo2001carteXI} and the development of a 12th-century notarial prosopography \cite{Modesti2012studi}. However, the transmission of these documents often suffers from material gaps, degradation of the writing medium, and textual omissions. These issues can limit the reconstruction opportunities offered by traditional diplomatic analysis.

These documents can be categorized according to their notarized content into the following two types: transactions and testamentary acts. The former constitute the majority of the corpus and include the following types:
\begin{itemize}
    \item deeds of sale (\textit{venditiones}), by which ownership of property was transferred in exchange for a monetary payment;
    \item donations (\textit{donationes}), recording the voluntary transfer of assets or rights;
    \item exchanges (\textit{permutationes}), documenting the reciprocal transfer of property between two parties;
    \item grants and concessions (\textit{concessiones}), through which the use, possession, or enjoyment of property or rights was assigned under specified conditions;
    \item emphyteutic leases (\textit{emphyteuses}), establishing long-term or even hereditary rights over land in return for a rent and obligations towards the granter;
    \item obligations (\textit{obligationes}), formalizing the assumption of financial or legal commitments by one or more parties;
    \item renunciations (\textit{renuntiationes}), recording the voluntary relinquishment of rights, claims, or legal entitlements.
\end{itemize}
The latter, testamentary acts, consist of wills, legal instruments through which individuals arrange the posthumous distribution of their property, appoint heirs, and establish pious or charitable bequests. Together, these documentary typologies express the principal legal instruments employed in the regulation of property transfers, contractual obligations, and inheritance within the medieval notarial tradition~\cite{Tamba1998instrumenti2}.

The transcription of these documents into digital form was carried out by professional diplomatists to ensure a high degree of accuracy and consistency in the representation of the original manuscripts. Owing to their long-term preservation, many of the manuscripts have suffered physical deterioration, resulting in lacunae within the text. Of the 1,184 documents, 725 contain at least one lacuna resulting in partial loss of text. When a lacuna is relatively short and its content can be inferred with a high degree of confidence from the surrounding context, or when it occurs in highly formulaic sections such as the protocol or eschatocol, paleographers may reconstruct the missing text with confidence. In such cases, the transcription explicitly marks both the presence of the damage and the proposed reconstructed text. In other instances, however, the missing text cannot be reliably recovered. When reconstruction is not possible, paleographers may estimate the character length of the missing text based on the physical dimensions of the damage and the size of the manuscript's script. The transcription then includes annotations indicating both the presence of the lacuna and the estimated number of missing characters. If even such an estimation cannot be made reliably, only the existence of the damage is recorded, without specifying its length. Table~\ref{tab_damages} reports the statistics of damages present in the dataset. As can be seen, in the majority of cases, the length of missing text can be estimated from the physical dimensions of the lacuna, and in more than half of these cases, the text can be inferred from the context. Lacunae of known length whose text cannot be inferred tend to be longer overall, with a mean of 12.41 characters, compared to 10.19 characters for lacunae where the text is inferred.

\input{tab_damages}

As highly structured records, notarial documents can be divided into three functional components: the protocol, the text body, and the eschatocol. There is also a special kind of notarial script called \textit{rogatio}, which is a summary note of the essential elements of the deed, drawn up by the notary in view of the subsequent drafting in extended form.  The protocol includes the opening formulas that establish the legal and chronological framework of the act. It generally opens with a verbal invocation (\textit{invocatio}), which in the earliest documents may also appear in symbolic form (e.g., a cross), followed by the chronological dating clause. From the eleventh century onward, the entire chronological dating is contained within the protocol, where it is typically expressed in complete form, combining the year of the Incarnation with additional elements, such as the indiction, the regnal year, or other dating systems employed in medieval documentary practice. Together, these elements provide the formal temporal framework within which the legal act was issued. The text body contains the main content of the legal act. This section records the parties' declarations, defines the object and conditions of the transaction, specifies the rights and obligations arising from the agreement, and frequently incorporates clauses concerning warranties, penalties, exceptions, and other legal provisions intended to govern the execution and validity of the act. In the end, the eschatocol constitutes the concluding section of the document. It generally includes the place of issue, the corroborative and authentication formulas, the references to witnesses, subscriptions, and, even if with many exceptions, the notarial sign. These elements certify the instrument's authenticity and attest to its legal validity. By contrast, \textit{rogationes} were not intended to function as legally authenticated documents in themselves. Rather, they served as working records preserved by the notary, summarizing the essential information required for the future compilation of the instrument. In many cases, these preliminary drafts constitute the only surviving evidence of transactions whose full notarial copies have been lost, making them an invaluable source for the study of medieval documentary practices~\cite{Costamagna1977charta}. 

Linguistically speaking, these texts differ considerably in their degree of formulaic patterning. The protocol and eschatocol are highly formulaic, relying on a relatively fixed repertoire of stock phrases and dating conventions that recur with limited variation across documents. The text body and the \textit{rogatio}, by contrast, are far less standardized, as their content is shaped by the specifics of each individual transaction and therefore varies considerably from document to document. Motivated by this structural and linguistic heterogeneity, we conduct a fine-grained analysis of language models' performance across the different parts of diplomatic texts, in order to assess whether their ability to restore missing content varies with the degree of formulaicity of each section. Figure~\ref{fig_1} shows the transcription of a notarial manuscript in the dataset together with the corresponding translation in English, divided into the four functional components. 


\input{fig_1}



\section{Experimental design \& Methodologies}
\label{design}
This section presents the different settings under which text restoration experiments are conducted, as well as the rationale followed during the experimental design phase. The objectives of this study are to explore and develop tools that can be implemented to automate or facilitate paleographers' reconstruction work on medieval notarial documents. Accordingly, the experimental design is intended to replicate real-world scenarios as closely as possible.

Two experimental settings are considered: the length-known setting and the length-unknown setting. In the former, the number of missing characters is assumed to be known through an estimate based on the physical dimensions of the lacuna, and this information is provided as input to the text restoration system. In the latter, the length of the missing text is assumed to be unavailable, and the system receives only the textual context surrounding the lacuna. The length-known setting is expected to yield more precise and controllable predictions, as the model is constrained by the estimated character count. The length-unknown setting is easier to implement in practice because it does not require manual estimation of the missing text length. It is also more broadly applicable, since the length of the missing text cannot always be reliably estimated. In our dataset, this is the case for 6.8\% of damages. Comparing these two settings also enables an evaluation of the trade-off between restoration accuracy and practical applicability.

\subsection{Dataset Partitions and Constructing the Test Set}
To guarantee the reliability of the test results, real lacunae present within the manuscripts are not adopted for evaluation purposes. Because there is no absolute historical guarantee that the text manually restored by paleographers matches the original verbatim, treating such reconstructions as ground truth introduces an empirical bias. Instead, the text restoration systems are evaluated using an artificially damaged test set built from intact textual regions of the manuscripts, ensuring a reliable ground truth.

The dataset is first divided into training set, development set and test set using a 8:1:1 ratio. Text samples are generated from each notarial document using a sliding-window. To simulate realistic damage scenarios on the intact regions of the test set, a synthetic damage pipeline is implemented to inject lacunae into each text sample according to Algorithm~\ref{alg:damage_injection} that is shown below.

\input{algorithm_decoding}

The length $l$ of the synthetic lacunae is modeled directly after the empirical distribution of real damages observed in the training set. These historical damage lengths are extracted from the training set from both lacunae with inferred text reconstructions and lacunae annotated with an estimated character length of the missing script. To eliminate anomalies and outliers arising from extreme physical degradation, an outer tail trim is applied to the collected lengths, keeping only the values between the 15th and 85th percentiles:
\begin{equation}
\mathcal{L}_{\text{pool}} = \{ l \in \mathcal{L}_{\text{train}} \mid P_{15} \le l \le P_{85} \}
\end{equation}

To thoroughly evaluate how robustly the restoration models scale with the size of the lacuna, the filtered length pool $\mathcal{L}_{\text{pool}}$ is divided at its median into two groups: a \textit{short pool} where lengths below or equal to the median and a \textit{long pool} where lengths above the median. In this study, the median length is 7 characters.

Given a target of $n$ damages per text window, the pipeline draws $\lceil n/2 \rceil$ spans from the short pool and $\lfloor n/2 \rfloor$ spans from the long pool. This balanced stratification isolates performance across varying degrees of contextual degradation and faithfully mirrors real-world paleographical damages. Each damage span is injected through a rejection-sampling procedure. A length is first drawn from the relevant pool and assigned a random starting position within the window. The resulting span is then checked against both the natural lacunae already present in the window and the spans injected in previous iterations. If any overlap is detected, the candidate is discarded and a new one is sampled. This process repeats until all $n$ damage spans have been successfully placed.

\subsection{Length-known Setting}
\label{Length-known Setting}
The length-known setting assumes that the length of the missing text can be reliably estimated from the size of the lacuna present in the manuscript, and this information is supplied as input together with the context surrounding the lacuna. This setting mirrors the real-world scenario encountered in the transcription work of paleographers, where the missing text is unknown but the size of the lacuna provides useful information. The language models chosen for this setting must satisfy the following criteria:
\begin{itemize}
    \item \textbf{Full control over the output length} The language model should be characterized by an architecture that allows control over the output length;
    \item \textbf{Support for the Latin language} The manuscripts under study are written in Latin, so the model must have sufficient exposure to Latin during training;
    \item \textbf{Open-source availability} For lanaguage models, Open-source models can be inspected, fine-tuned on the notarial corpus, and run locally, which is necessary both for adapting them to our work and for future implementations as tools for paleographers. For LLMs, only open-weight models are adopted as they guarantee reproducibility and long-term availability, independent of the commercial policies of third-party providers;
\end{itemize}

Following these principles, three systems are selected for the length-known setting: LaBERTa, PhilBERTa ~\cite{riemenschneider2023exploring}, and Aeneas~\cite{assael2025contextualizing}. The architecture and training details of these models have already been discussed in Section~\ref{lmforlatin} and are therefore not repeated here. Instead, the discussion focuses on the alignment between each model's tokenization scheme and the requirements of the length-known setting, since this property directly determines whether the output length can be effectively controlled. As for LLMs, Llama-70B Instruct~\cite{grattafiori2024llama3herdmodels} and Deepseek-v4-p~\cite{xu2026deepseek} are adopted.

\subsubsection{Aeneas}
\label{aeneas}

Aeneas~\cite{assael2025contextualizing} adopts a character-level tokenization in which every Latin letter, as well as each non-letter character, is treated as a single token. This scheme aligns naturally with the length-known setting: because there is a strict one-to-one correspondence between tokens and characters, constraining the length of the restored text reduces to constraining the number of tokens to be decoded. The model provides two dedicated placeholder symbols that distinguish the two restoration scenarios, where ``\texttt{-}'' marks a token from a lacuna whose length is known and ``\texttt{\#}'' marks a lacuna of unknown length. In the length-known setting, the inference relies exclusively on the former, supplying one placeholder per estimated missing character so that the model restores a span of exactly the desired size.

\subsubsection{LaBERTa and PhilBERTa}
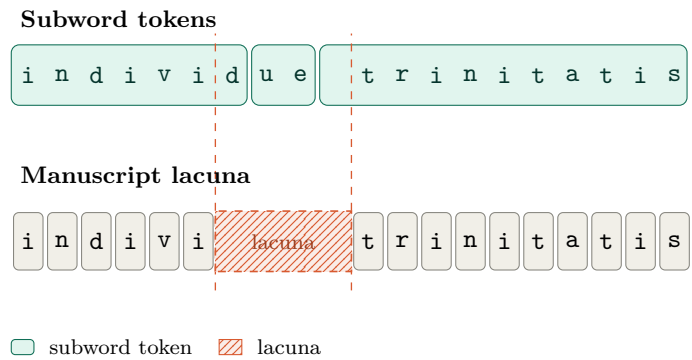
\begin{figure}[htbp]
    \centering
\input{fig_alignment_issue}
    \caption{An illustration of the misalignment between lacuna boundary and token boundary. The tokenizer in question is LaBERTa's tokenizer. In this case, the lacuna edges fall within tokens.}
    \label{fig_alignment}
\end{figure}

LaBERTa and PhilBERTa~\cite{riemenschneider2023exploring} have an encoder-based architecture, which allows the task to be formulated as masked language modeling (MLM), aligning naturally with the structure of the models. In MLM, the encoder outputs exactly one predicted token for each masked token in the input, allowing us to control the output length by manipulating the number of masks. However, like RoBERTa~\cite{zhuang-etal-2021-robustly}, both models employ a byte-pair encoding tokenizer, meaning that a single token may correspond to multiple characters, whereas the length constraint on the output is expressed in terms of the character count estimated from the lacuna size. As a result, there is no direct correspondence between the number of tokens to be decoded and the number of characters to be restored, and constraining the former does not reliably constrain the latter. A further issue introduced by the tokenizer is that the boundaries of a lacuna, as it occurs in a real manuscript, rarely coincide with the subword token boundaries learned by the model. A lacuna's boundaries may therefore fall in the middle of a token, splitting a learned subword unit into fragments that the model has rarely or never encountered during training. Figure~\ref{fig_alignment} illustrates such a case where the token learned by the model does not align with the lacuna boundary.

\begin{figure}[htbp]
    \centering
\input{fig_decoding}
    \caption{Text restoration workflow of LaBERTa and PhilBERTa models}
    \label{fig_decoding}
\end{figure}
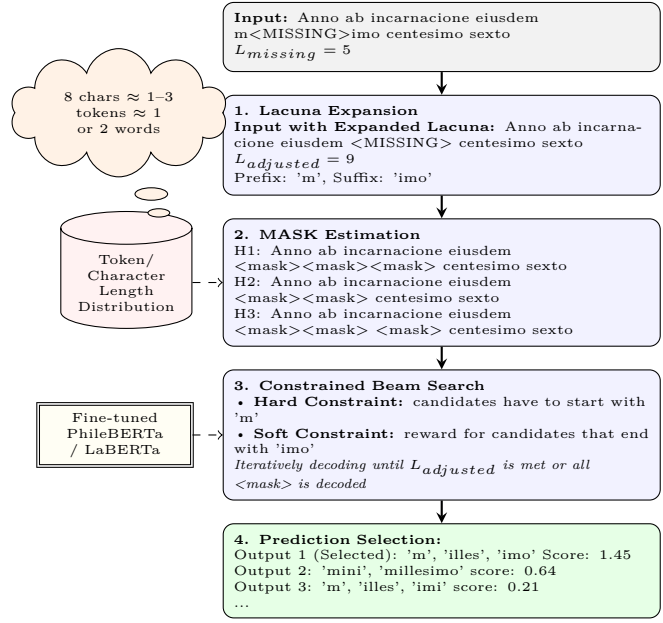

To resolve the boundary alignment issue and to constrain the output length effectively, a four-step decoding strategy is proposed. Figure~\ref{fig_decoding} illustrates the decoding strategy. The lacuna is first expanded to the closest token boundary on both sides to make sure that the lacuna boundary aligns with the token boundary learnt by the language models during the training phase. Then the token composition of expanded lacuna is estimated using the distribution observed from the training set and the character length of expanded lacuna. The decoding is conducted iteratively conditioned on the known prefix of the expanded lacuna and with extra bonus for a matching suffix until all expected tokens are decoded or the expected length is met.

\subsubsection{General-Purpose Large Language Model}
\label{general llms}
In addition to the above mentioned models, two general-purpose large language models (LLMs) are employed in a few-shot manner to investigate the feasibility of such tasks in a scenario with no training data.\

Llama-70B Instruct~\cite{grattafiori2024llama3herdmodels} is an open-weight instruction-tuned model based on Transformer architecture~\cite{NIPS2017_3f5ee243}. Deepseek-v4-pr~\cite{xu2026deepseek} is an open-weight Mixture-of-Experts LLM with 1.6 trillion parameters. This LLM is developed by DeepSeek AI and is known for its strong reasoning and multilingual capabilities. The prompt adopted for querying LLMs is demonstrated in Appendix~\ref{prompt}.


Word concordance information in the training set is also leveraged to further improve LLMs' performance. This approach draws inspiration from the manual text restoration procedures traditionally practiced by diplomatics experts, where restoration hypothesis are proposed by drawing on similar textual sources and the frequency patterns of words. This logic is adapted in this study by including in prompt word concordances derived from the training corpus, serving as a corpus-based analogue to the parallel passages and scribal habits an expert would otherwise consult by hand. For each lacuna, the word immediately preceding it, the word immediately following it, and, where the lacuna cuts into a word, the visible fragment (prefix or suffix) of that partially-obscured word are identified. Using these anchors, the training corpus is queried to extract the top three most frequent concordances along four dimensions: (1) words that both follow the preceding word and begin with the visible prefix; (2) words that both precede the following word and end with the visible suffix; (3) the most frequent words to follow the preceding word, regardless of fragment; and (4) the most frequent words to precede the following word, regardless of fragment.

\subsection{Length-unknown Setting}
\label{Length-unknown Setting}
The length-unknown setting assumes that the only available information for text restoration is the context text, while the character length of the missing portion remains unknown. Three language models were selected for this setting: LaTa, PhilTa~\cite{riemenschneider2023exploring}, and Aeneas~\cite{assael2025contextualizing}. In addition, two LLMs are tested like in the length-known setting: LLaMA-70B Instruct~\cite{grattafiori2024llama3herdmodels} and deepseek-v4-pro~\cite{xu2026deepseek}, evaluated in a few-shot manner.
\subsubsection{Aeneas}
\label{Aeneasunk}
As discussed in Section~\ref{aeneas}, Aeneas~\cite{assael2025contextualizing} is capable of performing restoration in scenarios where the missing text is unknown. In this case, the only difference lies in the use of ``\texttt{\#}'' to mark the lacuna, as opposed to ``\texttt{-}'', which masks individual letters. The system architecture and the initial weights remains unchanged from that used in the length-known setting.

\subsubsection{LaTa and PhilTa}


LaTa and PhilTa use the SentencePiece tokenizer~\cite{kudo-richardson-2018-sentencepiece}, which learns sub-word tokens. Consequently, model performance suffers when lacuna boundaries do not align with the token boundaries learned during training. Similar to the method illustrated in Figure~\ref{fig_decoding}, during inference the lacuna is first expanded to the nearest token boundary on both sides. Decoding is then conditioned on the known prefix of the expanded lacuna, with a bonus applied when the end of the prediction matches the known suffix. No restriction is imposed on the length of the output.

\subsubsection{General-Purpose Large Language Model}
LLMs are also evaluated in a length-unknown setting to assess their robustness. The prompt (see Figure~\ref{prompt}), temperature, and other hyperparameters remain the same. The only difference is that, in the length-known setting, the character length of the lacuna is provided.

\section{Results}
\label{sec:results}
All experiments reported were run five times, with the mean and 95\% confidence interval reported. For LLMs, this means that the same experiment was repeated five times. For models fine-tuned on the notarial manuscripts dataset, it means that five checkpoints were independently trained and evaluated, each using a different random seed.

As for evaluation metrics, character error rate (CER) and hit-rate (HR@N) are adopted. An additional metric named Overlap Score is introduced to provide a more direct conception of model's performance. This score is defined as following:

\begin{equation} \mathrm{overlap}(g, p) = \frac{\ell}{\lvert g\rvert + \lvert p\rvert - \ell}  
\qquad \ell = \bigl\lvert \operatorname{LCS}(g, p)\bigr\rvert 
\end{equation}

where $g$ and $p$ denote the ground-truth and predicted character strings, respectively, $\ell$ is the length of the longest common \emph{substring} shared by $g$ and $p$, i.e.\ the longest run of consecutive characters occurring in both. The score ranges from $0$, when the two strings share no common character run, to $1$, when they are identical.

\subsection{Length-known Setting}
Table~\ref{tab_known_by_length} presents the overall experimental results. The Aeneas model outperformed the other systems across all evaluation metrics, except for Overlap score and HR@10 in Long lacuna setting as it has some non-significant difference with the performance of LaBERTa. In general, model performance was higher for short lacunae and declined as lacuna length increased. DeepSeek achieved reasonably strong results despite lagging behind all fine-tuned models, which is expected. In contrast, Llama performed substantially worse, obtaining an HR@1 of only 5.62\%. 
Table~\ref{tab_known_divided} further breaks down these results by text type, revealing that, for supervised methods, the \textit{protocol} was comparatively the easiest section to restore, whereas the \textit{rogatio} proved the most challenging. Interestingly, DeepSeek achieves its best results on \textit{eschatocol} while results on \textit{protocol} and \textit{text body} are very close, while \textit{rogatio} turned out the be the most challenging component for DeepSeek as well. In general, multilingual PhilBERTa did not demonstrate significant advantage over its monolingual counterpart LaBERTa. These two models achieve very close results in all levels, with LaBERTa sometimes showing slight advantage over PhilBERTa.
\input{tab_known_by_length}
\input{tab_known_divided}

\subsection{Length-unknown Setting}
Tables~\ref{tab_unknown_by_length} and \ref{tab_unknown_divided} present the experimental results in the length-unknown setting. It is worth noting that the absence of information about the expected output length significantly increased the difficulty of text restoration: the best Hit-rate@1 dropped from 0.72 (short lacunae) and 0.29 (long lacunae) in the length-known setting to 0.42 and 0.18, respectively, in the length-unknown setting. Aeneas achieves the best results on short lacunae in terms of CER and HR@N, but lags behind LaTa on long lacunae, with a difference of 0.11 in CER and 0.12 in Hit-rate@1. PhilTa, the multilingual counterpart of LaTa, performs considerably worse in the length-unknown setting, consistently trailing LaTa and exhibiting substantially higher variance across runs. DeepSeek outperforms both LaTa and PhilTa on short lacunae, but falls behind LaTa on long lacunae in terms of Overlap score and HR@N. Notably, DeepSeek achieves the best CER and Overlap scores overall.

Regarding results divided by functional component, the patterns observed are coherent to those in the length-known setting: for fine-tuned models, the \textit{protocol} is relatively easier to restore, while the \textit{rogatio} is the hardest among all functional components. The performance of PhilTa falls considerably behind that of LaTa and has much higher variance across checkpoints, suggesting that multilinguality harms rather than helps model performance in this implementation. It is worth noticing that DeepSeek achieved very strong results on the \textit{eschatocol}: it outperformed all other systems with statistically significantly advantage in terms of CER, Overlap Score and HR@1.
\input{tab_unknown_by_length}
\input{tab_unknown_divided}

\subsection{Ablation experiments for LLM}
\input{tab_ablation_llm}
To further investigate the impact of prompting techniques on LLM output, additional ablation experiments were conducted by removing different parts from the prompt. Results are reported in Table~\ref{tab_deepseek_ablation}. They show that combining few-shot examples with concordance evidence consistently yields DeepSeek's best performance, in both the known-length and unknown-length settings, with non-overlapping confidence intervals confirming that the gains are robust. Few-shot prompting alone mainly improves CER, while concordance evidence alone mainly boosts hit rates, indicating that the two techniques address different weaknesses. When combined, they substantially improve the LLM's performance on all aspects.

\section{Discussion}
\label{sec:discussion}
\subsection{The effect of lacuna length on restoration difficulty}
\begin{figure}
    \centering
    \includegraphics[width=0.5\linewidth]{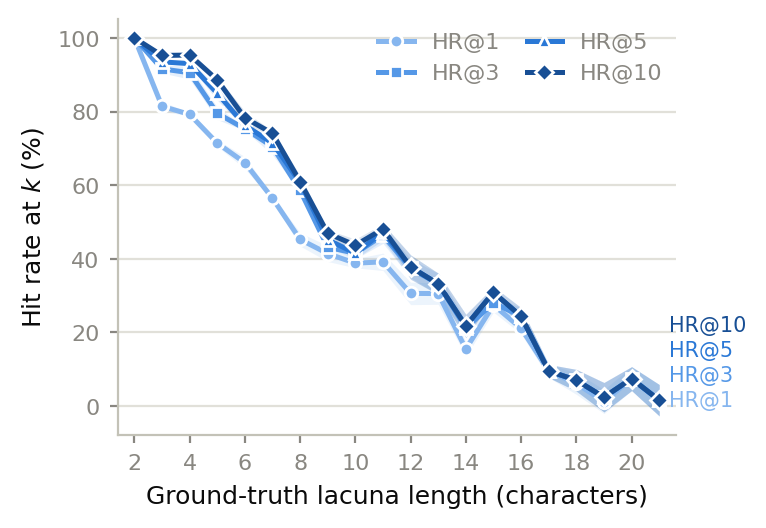}
    \caption{Relationship between Hit Rate and Lacuna Length of Aeneas in known-length setting}
    \label{fig_hitrate}
\end{figure}

In both the length-known and length-unknown settings, model performance is consistently better on short spans than on long spans. This is plausibly linked to the mechanics of decoding. As span length grows, the number of decoding steps required to generate a restoration increases, and each additional step introduces a further opportunity for the prediction to diverge from the ground truth. However, several other factors likely contribute as well. As lacuna length grows, the number of restorations that are semantically plausible but different from the ground truth is also likely to grow, independent of how the string is generated. Longer lacunae are also more likely to span additional content and remove the local contextual cues available for restoration. These explanations are not mutually exclusive and they cannot be cleanly separated using experimental design. Figure~\ref{fig_hitrate} demonstrates the relationship of Hit Rate and ground truth lacuna length of the Aeneas model's predictions. HR@1 declines approximately linearly with length until 20 characters, before flattening near the floor for the longest lacunae tested. HR@3, HR@5, and HR@10 follow the same overall shape and remain closely bunched with HR@1 throughout the length range. This demonstrates that the main reason of performance decay is likely not the existence of plausible restoration alternatives in a longer lacuna, as it would expect a widening between HR@10 and HR@1 with the growth of the lacuna length. The quasi-linear shape of the Hit Rate decline also suggests that an increasing number of decoding steps is unlikely to be the sole driver of the performance drop. Under a simple model where each step carries an independent risk of error, compounding errors would predict roughly exponential rather than linear decay.

\subsection{Formulaicity and restoration difficulty across functional components}
A consistent pattern that can be observed from Table~\ref{tab_known_divided} and Table~\ref{tab_unknown_divided} is that the two formulaic components, \textit{protocol} and \textit{eschatocol}, and \textit{text body}, are easier for every model than \textit{rogatio}. We attribute the observed phenomenon to two reasons. First, formulaic components are more frequent in the training material, while there is less \textit{text body} and \textit{rogatio}. Supervised models tend to perform better on the samples that appear frequently in training data. Among the 120 articles in the test set, we counted 104 \textit{protocol}, 107 \textit{text body}, 105 \textit{eschatocol}, but only 44 \textit{rogatio}, which may partially explain why fine-tuned models tend to achieve lower performance with \textit{rogatio}. Secondly, non-formulaic components exhibit greater lexical richness and more complex sentence structures, making the lacuna harder to restore.To investigate this further, we calculated the lexical density of the different document parts using the Type-Token Ratio (TTR)~\cite{laurs2024towards} and Hapax Legomena, here defined as words occurring only once within a given text type. As shown in Table~\ref{hapax}, \textit{rogatio} has both the highest TTR and the highest Hapax Legomena-to-token ratio among all functional components, confirming its high lexical richness. Given this distinct linguistic profile, one might expect the multilingual training of PhilBERTa and PhilTa to offer an advantage in such a lexically rich, non-formulaic context. Nevertheless, this expectation is not observed in our study, as PhilTa and PhilBERTa's performance is consistently lower than or on par with that of LaTa and LaBERTa.
 \input{tab_lexical_richness}
\subsection{Qualitative error analysis: what evaluation metrics miss}
Qualitative analysis was conducted on five notarial documents sampled from the test set by a professional paleographer, in order to further investigate the erroneous predictions made by the language models. It is expected that ``wrong'' predictions, despite differing from the ground truth, could still represent plausible alternatives to it, both semantically and grammatically. The evaluation relies on an objective ground truth, since the text was artificially masked. In real-world application, however, no such absolute ground truth exists, as the original text is already lost to the mists of time. Any prediction that fits the lacuna in terms of grammar, content, and length should be therefore considered a valid suggestion.

To facilitate and standardize the analysis process, paleographer was required to consider the grammaticality and semantic suitability of the predictions. These two aspects are defined as follows:
\begin{itemize}
    \item \textbf{Grammaticality} refers to possible gap-fillings that are consistent with Latin grammar at the morphological and syntactic level. For example, in a \textit{rogatio} written by the notary Angelo, we read: “Cartam vendicionis proprii et condicticii fecit Albertinus de Calcava, Morandus de Sole, Bonusmartinus de Martino de Berta de pecia que habent de Cararia iuxta viam et eos et iuxta fratrem presbiteri Rodaldi in no[Lacuna of 18 characters] prioris Sancti Victoris et eius fratrum ac successorum sub pena dupli et defensione”. \textit{(English: ``Albertinus de Calcava, Morandus de Sole, and Bonusmartinus de Martino de Berta made a charter of sale of their own property and agreed terms, concerning a plot of land which they hold from Cararia, next to the road and next to them, and next to the brother of the priest Rodaldus, in the [Lacuna] of the prior of St. Victor and his brothers and successors, under penalty of double payment and warranty of defense.'')}. The lacuna should be reconstructed as ``mine donno Martino'' \textit{(``in the name of lord Martino'')}, thus containing the name of a person. DeepSeek, however, also proposes in the lacuna length-unknown setting ``nime ecclesie'' \textit{(``in the name of the church'')} and ``nime monasterii'' \textit{(``in the name of the monastery'')} as possible reconstructions. Although neither of these alternatives is contextually correct, both are grammatically well-formed and fully compatible with the syntactic structure of the surrounding text.
    
    \item \textbf{Context} refers to possible gap-fillings that correctly capture the type of information required by the context, even when the proposed solution is grammatically incorrect. For example, in an emphyteutic lease, we read: ``Omnia qualiter super legitur presenti die do ego suprascriptus dominator predictum solum terre vob[Lacuna of 8 letters]ctis petitoribus vestrisque heredibus ad habendum, tenendum ac possidendum et faciendum quicquid vobis placuerit''. \textit{(English: ``All things as read above: on this present day I, the above-mentioned landlord, give the aforesaid plot of land [Lacuna] petitioners, and to your heirs, to have, hold, and possess, and to do whatever pleases you.'')}. The lacuna should be reconstructed as ``is iamdictis'' \textit{(``to you, the aforementioned'')}, referring back to the previously mentioned \textit{petitores} \textit{(petitioners)}. DeepSeek, however, both in the lacuna length-known and unknown setting, proposes a series of alternatives, such as ``is prenominatis iamdi'' \textit{(``to you, the aforenamed'')}, ``is memoratis iamdi'' \textit{(``to you, the aforementioned/recalled'')}, ``is sepedictis iamdi'' \textit{(``to you, the often-mentioned'')}, and ``is antedictis iamdi'' \textit{(``to you, the previously-said'')}. These reconstructions are not syntactically correct, since they result in a semantically redundant duplication of the reference to the previously mentioned persons. Nevertheless, they correctly identify the type of information required by the context. 
    
\end{itemize}

When a text restoration proposal does not align perfectly with the ground truth, but is nonetheless deemed grammatically acceptable and contextually fitting by a paleographer, this indicates that the proposal, despite being scored negatively by evaluation metrics, is actually a plausible alternative to the ground truth. For example, in an exchange, we read: “Ut neque nos qui supra suprascriptos conceditoris nostrisque heredibus ullam exinde abeatis molestacionem vel intencionem, sed predicta pecia terra vitata et aratoria uno se tenente qualiter super legitur promittimus nos suprascriptis Guilielmus et Berta germana sua et Masara et Rolandus et Aulivero et Fiopia germanis et germana conceditoris nostrisque [Lacuna of 6 characters]bus vobis suprascriptis Iohannes Bonus et Margarita iugalis acceptoris vestrisque heredibus omni tempore ab omni homine defensare et auctorizare”. \textit{(``So that neither we, the above-named grantors, nor our heirs, shall bring any trouble or claim against you concerning this; but we, the above-named Guilielmus and his sister Berta, and Masara, Rolandus, Aulivero, and Fiopia, siblings and grantors, and our [Lacuna], promise to defend and guarantee the aforesaid parcel of land, planted with vines and arable, held as one contiguous unit as described above, to you, the above-named spouses and recipients Iohannes Bonus and Margarita, and to your heirs, at all times, against everyone.'')}. The lacuna should be reconstructed as ``heredi'' \textit{(``heir(s)'')}. Alongside the correct solution, however, DeepSeek, in both the lacuna length-known and unkwown setting, proposes ``successori'' \textit{(``successor(s)'')}, and Aeneas, in the the lacuna length-known setting, ``fratri'' \textit{(``brother(s)'')}. Although none of these alternatives corresponds to the correct reading, they are all perfectly acceptable from both a grammatical and a contextual point of view, regardless of the actual size of the lacuna. This example is particularly significant because it shows that an incorrect reconstruction may nevertheless represent a meaningful and potentially useful hypothesis for the human scholar. 

Under lacuna length-known setting, qualitative analysis was conducted on the predictions of DeepSeek (few-shot + concordance), LaBERTa, and Aeneas. Under lacuna length-unknown setting, DeepSeek (few-shot + concordance), LaTa, and Aeneas. Analysis demonstrates that a considerable share of the predictions penalized by automatic metrics are in fact plausible restorations. In the lacuna length-known setting, the paleographer judged 23.6\% of DeepSeek's (few-shot + concordance) non-matching predictions, 18.8\% of LaBERTa's, and 13.3\% of Aeneas's to be plausible alternatives in terms of grammaticality and context. In the lacuna length-unknown setting, the corresponding proportions were 24.9\% for DeepSeek (few-shot + concordance), 20.3\% for LaTa, and 15.3\% for Aeneas. These results indicate that a non-trivial fraction of restorations flagged as errors, upon expert inspection, be judged acceptable, suggesting that automatic metrics alone tend to underestimate the true performance of the models.

Beyond these aggregate figures, the qualitative analysis also observed more fine-grained, model-specific tendencies. Aeneas, in the lacuna length-known setting, generally produced highly satisfactory restorations, both grammatically and contextually, when the gap involved proper names, kinship relations (e.g. \textit{filius} \textit{(``son'')}, \textit{uxor} \textit{(``wife'')}), or short lacunae. For example, in the aforementioned exchange written by Bonus Homo, we read: “Testes Lambertus clericus filius Petri clerici et Azo filius Petri de Petro de Azo et Azo filius Marinello et Martinus filius Iohanni Daretha et Fantinus filius Dominici corbelaro et Iohannes investitore filius [..8..]i Portunaro”. \textit{(``Witnesses: Lambertus the cleric, son of Petrus the cleric; and Azo, son of Petrus, of Petrus, of Azo; and Azo, son of Marinellus; and Martinus, son of Iohannes Daretha; and Fantinus, son of Dominicus the basket-maker; and Iohannes the guarantor/witness, son of [Lacuna] Portunarus.'')}. The lacuna contains the proper name ``Gierard''. In this case, Aeneas, when provided with the known length of the lacuna, is the only model to identify potentially correct solutions, such as ``Dominic'', ``Lambert'', and ``Rambert''. LaBERTa and LaTa, in turn, achieved similarly positive results when the missing text concerned common nouns or references to material goods, such as land, property, or other objects. Notably, LaBERTa was also observed to occasionally produce restorations that were acceptable both grammatically and contextually even in instances where all other models failed. For instance, in the aforementioned \textit{rogatio} written by Angelo, we read: “In predicto men[Lacuna] indicione predicta”. \textit{(English: ``In the aforementioned [Lacuna], in the aforementioned indiction'')}. The correct reconstruction of the lacuna is ``se'' \textit{(completing ``mense,'' i.e. ``month'')}, and LaBERTa is the only model that correctly identifies the presence of a syntactic pause between ``mense'' and ``indicione'', corresponding to a comma in the edition of the document, whereas all the other models fail to reproduce this feature.

\section{Conclusion}
\label{sec:conclusion}
This study investigated the feasibility of restoring missing text in damaged Latin notarial manuscripts using language models, taking into account both the practical needs of paleographers and the technical challenges of applying modern natural language processing methods to historical documents. By systematically comparing pre-trained language models of different architecture and general-purpose large language models across different real-world scenarios, we provide the first comprehensive evaluation of computational ancient text restoration. Our findings offer insights and methodological guidance that are relevant to text restoration in other historical and scholarly contexts.

The results demonstrate that text restoration of these documents cannot yet be fully automated, but can serve as a valuable assistive tool for paleographers. Model performance varied substantially depending on the functional component of the document being restored, with formulaic sections proving considerably easier to reconstruct than the non-formulaic sections. This finding has direct practical implications: computational restoration tools should be deployed selectively, with highest confidence in formulaic contexts and more cautious use in non-formulaic ones. Our comparison of length-known and length-unknown settings revealed a substantial performance gap, with overall hit-rate@1 dropping by 25.85\% when lacuna length became unavailable. This finding underscores the value of the paleographer's physical examination of the manuscript to estimate lacuna size from material evidence. However, the length-unknown setting remains important for cases where such estimation is not possible, and our results demonstrate that useful restoration hypotheses can still be generated in these cases. Regarding LLMs, an advantage of this approach is that it does not require fine-tuning the pre-trained model. When the length of missing text is known, LLM performance falls behind supervised methods, whereas when the length is unknown, the gap becomes non-significant. Nevertheless, this holds under the assumption that word concordance and few-shot examples can be derived from an existing dataset. When no such dataset is available and the LLM is queried via zero-shot prompting, hit-rate@1 drops significantly in both settings.

This work makes several contributions to the fields of digital humanities and natural language processing. First, it is the first study to systematically address the unknown-length scenario in manuscript restoration, a recurring challenge in real-world paleographic work. Second, it provides the first analysis of restoration performance across functional components of manuscripts, revealing the critical role of formulaicity in determining restoration difficulty. Third, it offers a practical evaluation of multiple model architectures and decoding strategies, providing guidance for future tool development. Finally, the fine-tuned models and methodologies developed here are made available to support paleographers in their work. Future work could explore multimodal models capable of estimating the length of missing text directly from visual input, as well as prompting strategies such as chain-of-thought reasoning~\cite{wei2022chain} to improve LLM restoration accuracy.
\hfill

\appendix
\section{Prompt}
\label{prompt}
\begin{figure}[H]
    \centering
\input{fig_prompt}
    \caption{Prompt used to query LLMs. The lacuna length information is provided only in length-known setting. The input text is truncated for limited space in paper.}
    \label{fig_prompt}
\end{figure}

\section*{Acknowledgments}
To be added

\section*{Disclosure statement:}
To be added

\printbibliography

\end{document}

%% file: tab_damages.tex
\begin{table}
    \centering
    \begin{tabular}{lccccc}
        \toprule
        Damage type            & Count   &Percentage       & Mean length & Median&Standard Deviation \\
        \midrule
        \textbf{Known length }          & 5807  &93.2\%& 11.61     &7  & 14.44  \\
        \quad Inferred text    & 3133     & 50.28\%   & 10.19   &6  & 16.56 \\
        \quad Estimated length & 2694   &   43.23\%    & 12.41    &9 & 11.47 \\
        \textbf{Unknown length}         & 424 &6.8\%       & /         & /   &/      \\
        \bottomrule
    \end{tabular}
    \caption{Statistics of textual damages in the manuscripts. \emph{Known length} aggregates the inferred-text and estimated-length cases. \emph{Mean length}, \emph{Median} and \emph{Standard Deviation} refer to character length and it may contain punctuations and spaces. ``/'' marks values that are undefined.}
    \label{tab_damages}
\end{table}

%% file: fig_1.tex
\begin{figure*}[t]%

    \tiny
    \hrule
    \vspace{1mm}
    \noindent \textbf{Protocol} \\
    \textit{Latin:} In nomine sancte et individue Trinitatis. Anno Domini millesimo centesimo trigesimo secundo, pridie kalendas aprelis, indicione decima.\\
    \textit{English:} In the name of the holy and indivisible Trinity. In the year of God one thousand one hundred and thirty-six, the day before the calends of April, tenth indiction.
    
    \vspace{2mm}
    \hrule
    \vspace{1mm}
    
    \noindent \textbf{Text Body} \\
    \textit{Latin:} Ego quidem Rainerius filius Lamberti de Beio hoc donacionis instrumento presenti die dono in honore Dei et ecclesie Sancti Victoris et tibi donno Alberio priori eiusdem ecclesie tuisque fratribus et successoribus proprium in perpetuum conducticium unde pertinuerit, id est omne quod ego habeo et teneo et michi pertinet iure vel actione a radice montis Sancti Victoris usque ad crucem de Dilvino ab Aposa usque ad rivum ex illa parte Barbiani et peciam unam terre aratorie in loco ubi dicitur Castellioni prope cruce de Piro cum ingressu et egressu suo usque in via publica et cum omnibus super se et infra se habentem in integrum. Finis vero eius: ab uno latere a sero et uno capite a meridie possidet Albertus de Rigiza, alio latere a mane detinet Rusticus de Emma terra de socru sua, alio capite ab aquilone adest via publica et si qui alii affines sunt; omnium quod infra hos fines michi pertinet in integrum pro remedio anime mee meeque uxoris nec non et patris et matris mee, in presenti dono et trado atque concedo supradicte ecclesie et tibi donno Alberio priori tuisque fratribus ac successoribus ad habendum, tenendum ac possidendum et quicquid tibi tuisque fratribus ac successoribus deinceps placuerit ad utilitatem eiusdem ecclesie faciendum. Ut nullam litem nullamque controversiam deinceps a me vel a meis heredibus quolibet modo aliquo in tempore vos vel vestri successores de cetero sustineatis ab omni quoque homine prescriptas res legittime defendere et auctorizare tibi et tuis heredibus promitto. Et si ego vel mei heredes prelibatam meam donationem sinplicem in totum vel pro parte audaci nisu quandoque infringere temtavero et eam semper inviolatam custodire noluero, penam triginta denariorum Lucensium libras tibi vel tuis successoribus dare promitto et insuper hanc donationem semper intactam conservare promitto. \\
    \textit{English:} I, Rainerius, son of Lambertus of Beio, in the present day, give by this document of donation in honor of God and of the church of Saint Victor and to you, father Alberius, prior of the aforesaid church, and to your brothers and successors, full property in perpetuity, wherever it should have been pertained, which is everything I own and possess and belongs to me, by right and action, from the foot of the mountain of Saint Victor as far as the cross of Dilvino, and from the Aposa as far as the river on that side of Barbianum, and moreover one plot of arable land in the place called Castellione, near the cross of Pero, with its right of entry and exit as far as the public road, and with everything it has above and below it, in its entirety. Moreover, its boundaries are: on the western and southern side, the land owned by Albertus of Rigiza, on the eastern side, held by Rusticus of Emma, the land of his mother-in-law, on the northern side, the public road, and all other boundaries, if there are any other. Everything that I own within these boundaries, in the present day, I give and deliver and grant in his integrity, for the salvation of my soul, and of my wife, and of my father and mother, to the aforesaid church and to father Alberius, prior, and to his brothers and successors, so that they may have, hold and possess it, and use until it will be of any utility for the aforesaid church. So that no arguing nor dispute hereafter from me or from my heirs, in any way or at any time, you or your successors shall have to endure, I promise to rightfully defend and authorize the aforesaid property against everyone for you and your heirs. And if I or my heirs shall at any time, by bold attempt, try to violate this my simple donation in whole or in part, or if I shall not wish to keep it always inviolate, I promise to pay to you or your successors the penalty of thirty denarii of Lucca, and moreover I promise to preserve this donation always intact. 
    
    \vspace{2mm}
    \hrule
    \vspace{1mm}
    
    \noindent \textbf{Eschatocol} \\
    \textit{Latin:} Actum in canonica Sancti Iohannis in Monte, indicione predicta. Prenominatus Rainerius hoc donacionis instrumentum ut supra legitur scribere rogavit. Iohannes presbiter de ecclesia Sancta Tecla, Lambertus investitor filius Petri de Leo, Lambertus de Auria, Petrus filius Alberti de Vivelinda, Grimaldus filius Bonifantini de Sancto Rofillo, Eldus de Verona rogati sunt testes. Gerardus tabellio hoc donacionis instrumentum ut supra legitur scripsi et firmavi.\\
    \textit{English:} Done in the canonry of Saint John on the Mount, in the aforesaid indiction.
The aforesaid Rainerius requested that this instrument of donation, as read above, may be written.
Iohannes, presbyter of the church of Saint Tecla, Lambertus investor, son of Petrus of Leo, Lambertus od Auria, Petrus, son of Albertus of Vivelinda, Grimaldus, son of Bonifantinus of Saint Rufillo, Eldus of Verona are called as witnesses.
I, Gerardus notary, wrote and signed this instrument of donation, as read before.

    \vspace{2mm}
    \hrule
    \vspace{1mm}
    
    \noindent \textbf{Rogationes} \\
    \textit{Latin:} Pridie kalendas aprelis, indicione x. Testis Iohannes presbiter de Sancta Tecla et Lambertus investitor filius Petri Leo et Lambertus de Auria et Petrus filius Alberti de Vivelinda et Eldus de Verona, Grimaldus filius Bonifantini de Sancto Rofillo. Cartulam donationis fecit Rainerius filius Lamberti de Beio pro remedio anime sue et de uxore sua et patris et matris sue in honore Dei et ecclesie Sancti Victoris et donno Alberio priori eiusdem ecclesie suisque fratribus ac successoribus de omnibus iuris et actionibus quod sibi pertinet a pede montium Sancti Victoris usque ad crucem Dilvini a Aposa usque ad rivum ex illa parte Barbiani et insuper peciam unam terre aratorie prope crucem de Pero in loco qui dicitur Castellioni sub pena et defensione. \\
    \textit{English:} The day before the calends of April, tenth indiction. Witnesses Iohannes, presbyter of Saint Tecla, and Lambertus investor, son of Petrus Leo, and Lambertus of Auria and Petrus, son of Albertus of Vivelinda and Eldus of Verona, Grimaldus, son of Bonifantinus of Saint Rufillo. Rainerius, son of Lambert of Beio, made a charter of donation for the salvation of his own soul, and of his wife, and of his father and mother, in honor of God and of the church of Saint Victor and of father Alberius, prior of the aforesaid church, and his brothers and successors, regarding all the rights and actions that belong to him from the foot of the mountain of Saint Victor as far as the cross of Dilvino, and from the Aposa as far as the river on that side of Barbianum, and moreover one plot of arable land near the cross of Pero, in the place called Castellione, under penalty and warranty.
    
    \vspace{1mm}
    \hrule
    \caption{A notarial document together with English translation. The document was produced in 1132, archive shelfmark \textit{Archivio di Stato di Bologna, Corporazioni religiose soppresse, S. Giovanni in Monte, 2/1342 n. 10a}}
    \label{fig_1}
\end{figure*}

%% file: algorithm_decoding.tex
\begin{algorithm}
\caption{Synthetic Damage Injection Pipeline}
\label{alg:damage_injection}
\begin{algorithmic}[1]
\REQUIRE Text sample $T$, target number of damages $n$, short pool $\mathcal{L}_{\text{short}}$, long pool $\mathcal{L}_{\text{long}}$
\STATE $\mathcal{D} \gets \emptyset$ \COMMENT{Initialize set of injected damage spans}
\FORALL{$(\mathcal{P}, k) \in \{(\mathcal{L}_{\text{short}}, \lceil n/2 \rceil),\ (\mathcal{L}_{\text{long}}, \lfloor n/2 \rfloor)\}$}
    \STATE $c \gets 0$ \COMMENT{Spans injected from this pool}
    \WHILE{$c < k$}
        \STATE $l \gets \text{sample\_length}(\mathcal{P})$ \COMMENT{Sample a length from pool $\mathcal{P}$}
        \STATE $s \gets \text{random\_position}(T, l)$ \COMMENT{Pick a random starting position}
        \STATE $e \gets s + l$
        \IF{$[s, e)$ overlaps with any natural lacuna in $T$ \OR $[s, e)$ overlaps with any already selected span in $\mathcal{D}$}
            \STATE \textbf{continue} \COMMENT{Reject candidate span and restart}
        \ELSE
            \STATE $\mathcal{D} \gets \mathcal{D} \cup \{[s, e)\}$ \COMMENT{Accept and inject the lacuna}
            \STATE $c \gets c + 1$
        \ENDIF
    \ENDWHILE
\ENDFOR
\end{algorithmic}
\end{algorithm}

%% file: fig_alignment_issue.tex
\begin{tikzpicture}[font=\small]

\def\CW{0.45} 

\node[anchor=west, font=\small\bfseries] at (0,3.35)
  {Subword tokens};

\draw[rounded corners=3pt, fill=tokenfill, draw=tokenstroke, line width=0.4pt]
  (0,2.2) rectangle (3.12,3.0);
\draw[rounded corners=3pt, fill=tokenfill, draw=tokenstroke, line width=0.4pt]
  (3.18,2.2) rectangle (4.02,3.0);
\draw[rounded corners=3pt, fill=tokenfill, draw=tokenstroke, line width=0.4pt]
  (4.08,2.2) rectangle (8.94,3.0);

\foreach \i/\c in {
0/i,1/n,2/d,3/i,4/v,5/i,6/d,7/u,8/e,
10/t,11/r,12/i,13/n,14/i,15/t,16/a,17/t,18/i,19/s}
  \node[text=tokentext, font=\ttfamily] at ({\i*\CW+\CW/2},2.6) {\c};

\draw[coral, dashed, line width=0.5pt] (2.70,-0.25) -- (2.70,3.15);
\draw[coral, dashed, line width=0.5pt] (4.50,-0.25) -- (4.50,3.15);

\node[anchor=west, font=\small\bfseries] at (0,1.25) {Manuscript lacuna};

\foreach \i/\c in {0/i,1/n,2/d,3/i,4/v,5/i} {
  \draw[rounded corners=2pt, fill=cellfill, draw=cellstroke, line width=0.4pt]
    ({\i*\CW+0.03},0.02) rectangle ({(\i+1)*\CW-0.03},0.78);
  \node[font=\ttfamily] at ({\i*\CW+\CW/2},0.4) {\c};
}

\fill[coralfill] (2.70,0) rectangle (4.50,0.8);
\fill[pattern=north east lines, pattern color=coral] (2.70,0) rectangle (4.50,0.8);
\draw[coral, dashed, line width=0.6pt] (2.70,0) rectangle (4.50,0.8);
\node[text=coraltext, font=\footnotesize] at (3.60,0.4) {lacuna};

\foreach \i/\c in {10/t,11/r,12/i,13/n,14/i,15/t,16/a,17/t,18/i,19/s} {
  \draw[rounded corners=2pt, fill=cellfill, draw=cellstroke, line width=0.4pt]
    ({\i*\CW+0.03},0.02) rectangle ({(\i+1)*\CW-0.03},0.78);
  \node[font=\ttfamily] at ({\i*\CW+\CW/2},0.4) {\c};
}


\draw[rounded corners=1.5pt, fill=tokenfill, draw=tokenstroke, line width=0.4pt]
  (0,-1.1) rectangle (0.3,-0.9);
\node[anchor=west, font=\footnotesize] at (0.38,-1.0) {subword token};

\fill[coralfill] (2.75,-1.1) rectangle (3.05,-0.9);
\fill[pattern=north east lines, pattern color=coral] (2.75,-1.1) rectangle (3.05,-0.9);
\draw[coral, line width=0.4pt] (2.75,-1.1) rectangle (3.05,-0.9);
\node[anchor=west, font=\footnotesize] at (3.13,-1.0) {lacuna};


\end{tikzpicture}

%% file: fig_decoding.tex
\usetikzlibrary{shapes.geometric}

    \begin{tikzpicture}[
        node distance=0.3cm,
        every node/.style={font=\tiny},
        block/.style={rectangle, draw, fill=blue!5, text width=5.5cm, align=left, rounded corners, minimum height=0.6cm, inner sep=4pt},
        resource/.style={cylinder, draw, shape border rotate=90, aspect=0.25, fill=red!5, text width=1.5cm, align=center, font=\tiny},
        model/.style={rectangle, draw, double, fill=yellow!5, text width=1.8cm, align=center},
        arrow/.style={thick,->,>=stealth},
        note/.style={cloud callout, draw, fill=orange!10, callout relative pointer={(0.2,-0.5)}, aspect=2.5, font=\tiny, text width=1.7cm, align=center}
    ]

    \node (start) [block, fill=gray!10] {\textbf{Input:} Anno ab incarnacione eiusdem m<MISSING>imo centesimo sexto \\ $L_{missing}=5$};
    
    \node (s1) [block, below=of start] {\textbf{1. Lacuna Expansion} \\\textbf{Input with Expanded Lacuna:} Anno ab incarnacione eiusdem <MISSING> centesimo sexto \\ $L_{adjusted}=9$\\
    Prefix: 'm', Suffix: 'imo'};
    
    \node (s2) [block, below=of s1] {
        \textbf{2. MASK Estimation} \\
        \tiny
        H1: Anno ab incarnacione eiusdem <mask><mask><mask> centesimo sexto \\
        H2: Anno ab incarnacione eiusdem <mask><mask> centesimo sexto \\
        H3: Anno ab incarnacione eiusdem <mask><mask> <mask> centesimo sexto
    };
    
    \node (dist) [resource, left=of s2, xshift=-0.1cm] {Token/ Character Length Distribution};
    \node at (dist.north) [note, yshift=1.3cm, xshift=-0.1cm] {8 chars $\approx$ 1--3 tokens $\approx$ 1 or 2 words};
    
\node (s3) [block, below=of s2] {
        \textbf{3. Constrained Beam Search} \\
        \tiny
        \textbullet\ \textbf{Hard Constraint:} candidates have to start with 'm' \\
        \textbullet\ \textbf{Soft Constraint:} reward for candidates that end with 'imo' \\
        \textit{Iteratively decoding until $L_{adjusted}$ is met or all <mask> is decoded}
    };

\node (roberta) [model, left=of s3, xshift=-0.1cm] {Fine-tuned PhileBERTa / LaBERTa};

    \node (final) [block, fill=green!10, below=of s3] {\textbf{4. Prediction Selection:} \\
        Output 1 (Selected): 'm', 'illes', 'imo' Score: 1.45  \\
        Output 2: 'mini', 'millesimo' score: 0.64\\
        Output 3: 'm', 'illes', 'imi' score: 0.21\\...};

    \draw [arrow] (start) -- (s1);
    \draw [arrow] (s1) -- (s2);
    \draw [dashed, ->] (dist) -- (s2);
    \draw [arrow] (s2) -- (s3);
    \draw [dashed, ->] (roberta) -- (s3);
    \draw [arrow] (s3) -- (final);

    \end{tikzpicture}

%% file: tab_known_by_length.tex
\begin{table}[h]
\centering

\resizebox{0.95\textwidth}{!}{
\begin{tabular}{ll|cccccc}
\hline
Lacuna Length & Model & CER & Overlap & HR@1 & HR@3 & HR@5 & HR@10 \\ \hline
Overall & Aeneas & $\mathbf{\underline{30.26_{\pm 0.20}}}$ & $\mathbf{\underline{60.79_{\pm 0.42}}}$ & $\mathbf{\underline{50.71_{\pm 0.54}}}$ & $\mathbf{\underline{58.08_{\pm 0.42}}}$ & $\mathbf{\underline{59.69_{\pm 0.38}}}$ & $\mathbf{\underline{61.20_{\pm 0.29}}}$ \\
 & LaBERTa & $59.77_{\pm 1.54}$ & $53.94_{\pm 0.96}$ & $32.95_{\pm 0.64}$ & $40.66_{\pm 0.49}$ & $43.03_{\pm 0.24}$ & $45.17_{\pm 0.54}$ \\
 & PhilBERTa & $61.59_{\pm 1.31}$ & $52.70_{\pm 0.57}$ & $31.05_{\pm 0.67}$ & $38.97_{\pm 0.72}$ & $41.97_{\pm 0.73}$ & $44.86_{\pm 1.18}$ \\
 & DeepSeek & $47.55_{\pm 0.58}$ & $53.00_{\pm 0.86}$ & $26.38_{\pm 1.12}$ & $31.98_{\pm 1.13}$ & $33.24_{\pm 1.24}$ & $34.28_{\pm 1.71}$ \\
 & Llama & $70.78_{\pm 0.76}$ & $33.82_{\pm 0.20}$ & $5.62_{\pm 0.49}$ & $8.08_{\pm 0.28}$ & $10.27_{\pm 0.36}$ & $12.22_{\pm 0.42}$ \\ \hline
Short & Aeneas & $\mathbf{\underline{18.16_{\pm 0.27}}}$ & $\mathbf{\underline{78.43_{\pm 0.41}}}$ & $\mathbf{\underline{72.34_{\pm 0.54}}}$ & $\mathbf{\underline{82.54_{\pm 0.58}}}$ & $\mathbf{\underline{84.95_{\pm 0.43}}}$ & $\mathbf{\underline{87.39_{\pm 0.44}}}$ \\
 & LaBERTa & $64.23_{\pm 2.01}$ & $62.29_{\pm 0.99}$ & $43.42_{\pm 1.11}$ & $51.22_{\pm 0.48}$ & $53.49_{\pm 0.38}$ & $55.08_{\pm 0.21}$ \\
 & PhilBERTa & $66.64_{\pm 1.89}$ & $61.00_{\pm 0.18}$ & $41.39_{\pm 0.86}$ & $50.81_{\pm 0.58}$ & $53.63_{\pm 0.81}$ & $56.58_{\pm 1.51}$ \\
 & DeepSeek & $44.87_{\pm 1.21}$ & $66.86_{\pm 0.23}$ & $38.67_{\pm 0.43}$ & $47.27_{\pm 0.62}$ & $48.66_{\pm 0.79}$ & $49.61_{\pm 1.10}$ \\
 & Llama & $70.97_{\pm 1.23}$ & $43.64_{\pm 0.37}$ & $9.77_{\pm 0.74}$ & $13.61_{\pm 0.44}$ & $17.47_{\pm 0.62}$ & $20.26_{\pm 0.40}$ \\ \hline
Long & Aeneas & $\mathbf{\underline{42.35_{\pm 0.32}}}$ & $43.16_{\pm 0.51}$ & $\mathbf{\underline{29.08_{\pm 0.57}}}$ & $\mathbf{\underline{33.63_{\pm 0.44}}}$ & $\mathbf{\underline{34.44_{\pm 0.50}}}$ & $35.02_{\pm 0.32}$ \\
 & LaBERTa & $55.30_{\pm 1.65}$ & $\underline{45.58_{\pm 1.12}}$ & $22.47_{\pm 1.24}$ & $30.10_{\pm 0.68}$ & $32.58_{\pm 0.35}$ & $\underline{35.25_{\pm 0.87}}$ \\
 & PhilBERTa & $56.55_{\pm 1.14}$ & $44.41_{\pm 1.27}$ & $20.71_{\pm 1.45}$ & $27.12_{\pm 0.95}$ & $30.31_{\pm 1.00}$ & $33.15_{\pm 1.28}$ \\
 & DeepSeek & $50.24_{\pm 1.81}$ & $39.13_{\pm 1.75}$ & $14.08_{\pm 1.96}$ & $16.68_{\pm 1.81}$ & $17.81_{\pm 1.88}$ & $18.93_{\pm 2.39}$ \\
 & Llama & $70.60_{\pm 0.39}$ & $23.99_{\pm 0.08}$ & $1.46_{\pm 0.24}$ & $2.55_{\pm 0.26}$ & $3.06_{\pm 0.40}$ & $4.18_{\pm 0.60}$ \\ \hline
\end{tabular}
}

\caption{Known-length setting results (mean $\pm$ 95\% confidence interval over 5 checkpoints).}
\label{tab_known_by_length}
\end{table}

%% file: tab_known_divided.tex
\begin{table}[h]
\centering

\resizebox{0.95\textwidth}{!}{
\begin{tabular}{ll|cccccc}
\hline
Component & Model & CER & Overlap & HR@1 & HR@3 & HR@5 & HR@10 \\ \hline
Protocol & Aeneas & $\mathbf{\underline{13.40_{\pm 2.38}}}$ & $\mathbf{\underline{77.42_{\pm 2.93}}}$ & $\mathbf{\underline{68.36_{\pm 4.02}}}$ & $\mathbf{\underline{73.73_{\pm 2.11}}}$ & $\mathbf{\underline{74.93_{\pm 2.42}}}$ & $\mathbf{\underline{75.22_{\pm 2.11}}}$ \\
 & LaBERTa & $36.82_{\pm 0.91}$ & $67.51_{\pm 1.27}$ & $46.57_{\pm 2.42}$ & $56.42_{\pm 1.55}$ & $58.21_{\pm 1.85}$ & $60.60_{\pm 1.66}$ \\
 & PhilBERTa & $39.19_{\pm 5.69}$ & $63.49_{\pm 2.03}$ & $40.30_{\pm 2.62}$ & $51.64_{\pm 2.49}$ & $54.93_{\pm 4.02}$ & $59.10_{\pm 5.50}$ \\
 & DeepSeek & $51.11_{\pm 11.08}$ & $51.96_{\pm 5.39}$ & $24.40_{\pm 0.90}$ & $32.87_{\pm 2.19}$ & $34.06_{\pm 2.97}$ & $36.15_{\pm 4.54}$ \\
 & Llama & $60.38_{\pm 2.55}$ & $44.88_{\pm 1.13}$ & $8.96_{\pm 2.27}$ & $13.73_{\pm 3.05}$ & $17.91_{\pm 3.47}$ & $18.81_{\pm 2.81}$ \\ \hline
Text Body & Aeneas & $\mathbf{\underline{30.64_{\pm 0.39}}}$ & $\mathbf{\underline{60.46_{\pm 0.57}}}$ & $\mathbf{\underline{50.11_{\pm 0.66}}}$ & $\mathbf{\underline{58.31_{\pm 0.48}}}$ & $\mathbf{\underline{59.78_{\pm 0.44}}}$ & $\mathbf{\underline{61.37_{\pm 0.30}}}$ \\
 & LaBERTa & $59.64_{\pm 1.83}$ & $55.35_{\pm 1.09}$ & $33.35_{\pm 1.27}$ & $41.30_{\pm 0.50}$ & $43.81_{\pm 0.41}$ & $45.95_{\pm 0.68}$ \\
 & PhilBERTa & $60.65_{\pm 1.94}$ & $55.08_{\pm 0.79}$ & $32.50_{\pm 0.78}$ & $40.27_{\pm 1.07}$ & $43.19_{\pm 0.90}$ & $46.05_{\pm 1.34}$ \\
 & DeepSeek & $46.78_{\pm 1.18}$ & $53.61_{\pm 1.33}$ & $26.31_{\pm 1.30}$ & $31.75_{\pm 1.63}$ & $32.90_{\pm 1.52}$ & $33.97_{\pm 2.08}$ \\
 & Llama & $71.27_{\pm 1.13}$ & $33.58_{\pm 0.37}$ & $5.12_{\pm 0.58}$ & $7.21_{\pm 0.72}$ & $8.84_{\pm 0.77}$ & $10.93_{\pm 0.81}$ \\ \hline
Eschatocol & Aeneas & $\mathbf{\underline{28.15_{\pm 0.33}}}$ & $\mathbf{\underline{62.22_{\pm 0.37}}}$ & $\mathbf{\underline{52.66_{\pm 0.52}}}$ & $\mathbf{\underline{58.29_{\pm 0.99}}}$ & $\mathbf{\underline{59.70_{\pm 1.20}}}$ & $\mathbf{\underline{61.11_{\pm 0.71}}}$ \\
 & LaBERTa & $64.51_{\pm 2.90}$ & $49.79_{\pm 1.51}$ & $31.66_{\pm 1.59}$ & $38.79_{\pm 1.67}$ & $41.21_{\pm 0.44}$ & $42.71_{\pm 0.88}$ \\
 & PhilBERTa & $69.00_{\pm 3.59}$ & $47.00_{\pm 1.75}$ & $28.14_{\pm 1.82}$ & $34.37_{\pm 1.80}$ & $37.69_{\pm 1.46}$ & $40.30_{\pm 1.20}$ \\
 & DeepSeek & $43.24_{\pm 2.59}$ & $58.82_{\pm 1.24}$ & $33.67_{\pm 0.98}$ & $39.78_{\pm 0.63}$ & $41.88_{\pm 0.58}$ & $41.99_{\pm 0.86}$ \\
 & Llama & $66.66_{\pm 0.95}$ & $36.55_{\pm 0.92}$ & $7.85_{\pm 0.93}$ & $11.77_{\pm 0.93}$ & $16.30_{\pm 1.02}$ & $18.51_{\pm 1.59}$ \\ \hline
Rogatio & Aeneas & $\mathbf{\underline{41.41_{\pm 1.95}}}$ & $\mathbf{\underline{50.60_{\pm 1.68}}}$ & $\mathbf{\underline{40.90_{\pm 2.03}}}$ & $\mathbf{\underline{46.67_{\pm 0.50}}}$ & $\mathbf{\underline{49.91_{\pm 1.00}}}$ & $\mathbf{\underline{51.71_{\pm 1.00}}}$ \\
 & LaBERTa & $66.01_{\pm 2.77}$ & $42.94_{\pm 2.11}$ & $24.14_{\pm 3.30}$ & $29.91_{\pm 2.00}$ & $31.53_{\pm 1.37}$ & $34.59_{\pm 1.28}$ \\
 & PhilBERTa & $68.69_{\pm 1.03}$ & $39.21_{\pm 1.03}$ & $20.18_{\pm 0.61}$ & $30.09_{\pm 2.32}$ & $32.97_{\pm 2.58}$ & $35.86_{\pm 0.94}$ \\
 & DeepSeek & $59.79_{\pm 3.45}$ & $38.18_{\pm 0.74}$ & $14.79_{\pm 2.15}$ & $18.84_{\pm 0.96}$ & $19.39_{\pm 1.21}$ & $21.22_{\pm 1.37}$ \\
 & Llama & $80.95_{\pm 1.65}$ & $24.03_{\pm 0.64}$ & $3.24_{\pm 0.61}$ & $4.32_{\pm 0.94}$ & $5.23_{\pm 1.66}$ & $6.31_{\pm 1.77}$ \\ \hline
\end{tabular}
}
\caption{Known-length setting results divided by the functional components of notarial manuscripts(mean $\pm$ 95\% confidence interval over 5 checkpoints).}
\label{tab_known_divided}
\end{table}

%% file: tab_unknown_by_length.tex
\begin{table}[h]
\centering
\resizebox{0.95\textwidth}{!}{
\begin{tabular}{ll|cccccc}
\hline
Lacuna Length & Model & CER & Overlap & HR@1 & HR@3 & HR@5 & HR@10 \\ \hline
Overall & Aeneas & $70.33_{\pm 1.39}$ & $41.49_{\pm 0.51}$ & $24.05_{\pm 0.65}$ & $\mathbf{\underline{37.61_{\pm 0.67}}}$ & $\mathbf{\underline{40.66_{\pm 0.25}}}$ & $\mathbf{\underline{43.85_{\pm 0.49}}}$ \\
 & LaTa & $72.72_{\pm 2.27}$ & $47.71_{\pm 0.60}$ & $\underline{24.86_{\pm 0.96}}$ & $30.86_{\pm 1.71}$ & $32.07_{\pm 2.44}$ & $32.66_{\pm 3.03}$ \\
 & PhilTa & $130.29_{\pm 27.24}$ & $28.71_{\pm 2.49}$ & $11.20_{\pm 1.48}$ & $13.86_{\pm 2.05}$ & $14.20_{\pm 2.31}$ & $14.31_{\pm 2.41}$ \\
 & DeepSeek & $\mathbf{\underline{67.15_{\pm 1.59}}}$ & $\mathbf{\underline{49.94_{\pm 0.09}}}$ & $23.86_{\pm 0.24}$ & $28.41_{\pm 0.22}$ & $30.04_{\pm 0.13}$ & $30.76_{\pm 0.20}$ \\
 & Llama & $81.18_{\pm 0.94}$ & $30.72_{\pm 0.31}$ & $3.48_{\pm 0.16}$ & $6.08_{\pm 0.40}$ & $7.54_{\pm 0.42}$ & $8.90_{\pm 0.54}$ \\ \hline
Short & Aeneas & $\underline{69.83_{\pm 2.32}}$ & $56.87_{\pm 0.44}$ & $\mathbf{\underline{41.97_{\pm 0.72}}}$ & $\mathbf{\underline{59.19_{\pm 1.38}}}$ & $\mathbf{\underline{63.83_{\pm 0.35}}}$ & $\mathbf{\underline{69.15_{\pm 0.88}}}$ \\
 & LaTa & $85.60_{\pm 5.20}$ & $51.82_{\pm 0.90}$ & $31.32_{\pm 1.13}$ & $37.73_{\pm 1.84}$ & $38.81_{\pm 2.84}$ & $39.12_{\pm 3.33}$ \\
 & PhilTa & $142.72_{\pm 40.93}$ & $32.72_{\pm 2.93}$ & $15.80_{\pm 2.13}$ & $18.85_{\pm 3.18}$ & $19.12_{\pm 3.28}$ & $19.15_{\pm 3.28}$ \\
 & DeepSeek & $74.83_{\pm 3.65}$ & $\mathbf{\underline{62.52_{\pm 0.10}}}$ & $35.18_{\pm 0.33}$ & $41.14_{\pm 0.32}$ & $42.85_{\pm 0.37}$ & $43.12_{\pm 0.52}$ \\
 & Llama & $87.09_{\pm 1.80}$ & $39.01_{\pm 0.44}$ & $5.91_{\pm 0.27}$ & $10.56_{\pm 0.74}$ & $13.00_{\pm 0.77}$ & $15.21_{\pm 0.93}$ \\ \hline
Long & Aeneas & $70.84_{\pm 0.86}$ & $26.12_{\pm 0.61}$ & $6.14_{\pm 0.67}$ & $16.03_{\pm 0.44}$ & $17.49_{\pm 0.58}$ & $18.54_{\pm 0.35}$ \\
 & LaTa & $59.84_{\pm 1.03}$ & $\mathbf{\underline{43.61_{\pm 1.23}}}$ & $\mathbf{\underline{18.41_{\pm 0.88}}}$ & $\mathbf{\underline{24.00_{\pm 1.63}}}$ & $\mathbf{\underline{25.32_{\pm 2.12}}}$ & $\mathbf{\underline{26.20_{\pm 2.79}}}$ \\
 & PhilTa & $117.86_{\pm 17.91}$ & $24.71_{\pm 2.05}$ & $6.61_{\pm 1.03}$ & $8.88_{\pm 1.08}$ & $9.29_{\pm 1.44}$ & $9.46_{\pm 1.65}$ \\
 & DeepSeek & $\underline{59.51_{\pm 0.59}}$ & $37.41_{\pm 0.11}$ & $12.58_{\pm 0.17}$ & $15.73_{\pm 0.23}$ & $17.29_{\pm 0.26}$ & $18.45_{\pm 0.31}$ \\
 & Llama & $75.27_{\pm 0.15}$ & $22.42_{\pm 0.21}$ & $1.05_{\pm 0.09}$ & $1.60_{\pm 0.12}$ & $2.07_{\pm 0.18}$ & $2.58_{\pm 0.43}$ \\ \hline
\end{tabular}
}
\caption{Unknown-length setting: results divided by text length (mean $\pm$ 95\% CI, n=5 seeds).}
\label{tab_unknown_by_length}
\end{table}

%% file: tab_unknown_divided.tex
\begin{table}[h]
\centering
\resizebox{0.95\textwidth}{!}{
\begin{tabular}{ll|cccccc}
\hline
Component & Model & CER & Overlap & HR@1 & HR@3 & HR@5 & HR@10 \\ \hline
Protocol & Aeneas & $\mathbf{\underline{39.45_{\pm 1.98}}}$ & $56.94_{\pm 0.73}$ & $37.91_{\pm 2.11}$ & $\underline{51.34_{\pm 2.11}}$ & $\underline{51.94_{\pm 2.42}}$ & $\underline{53.13_{\pm 2.11}}$ \\
 & LaTa & $48.84_{\pm 5.42}$ & $\underline{58.77_{\pm 1.21}}$ & $\underline{40.90_{\pm 3.84}}$ & $47.16_{\pm 2.81}$ & $48.96_{\pm 2.42}$ & $50.15_{\pm 2.81}$ \\
 & PhilTa & $65.33_{\pm 16.68}$ & $46.19_{\pm 3.27}$ & $25.67_{\pm 6.06}$ & $35.22_{\pm 3.84}$ & $35.52_{\pm 4.23}$ & $36.12_{\pm 4.80}$ \\
 & DeepSeek & $71.49_{\pm 6.37}$ & $52.24_{\pm 0.89}$ & $20.00_{\pm 1.66}$ & $25.67_{\pm 1.55}$ & $25.97_{\pm 1.66}$ & $28.06_{\pm 2.42}$ \\
 & Llama & $64.11_{\pm 0.28}$ & $40.57_{\pm 0.67}$ & $3.58_{\pm 1.02}$ & $10.75_{\pm 0.83}$ & $12.24_{\pm 0.83}$ & $14.03_{\pm 1.02}$ \\ \hline
Text Body & Aeneas & $72.00_{\pm 1.40}$ & $40.44_{\pm 0.57}$ & $22.76_{\pm 0.78}$ & $\underline{36.61_{\pm 0.70}}$ & $\mathbf{\underline{40.15_{\pm 0.39}}}$ & $\mathbf{\underline{43.46_{\pm 0.61}}}$ \\
 & LaTa & $68.99_{\pm 5.18}$ & $\mathbf{\underline{51.79_{\pm 0.78}}}$ & $\mathbf{\underline{27.25_{\pm 1.48}}}$ & $33.72_{\pm 2.38}$ & $34.74_{\pm 2.94}$ & $35.32_{\pm 3.46}$ \\
 & PhilTa & $118.90_{\pm 19.50}$ & $28.98_{\pm 2.83}$ & $11.38_{\pm 1.53}$ & $13.95_{\pm 2.58}$ & $14.25_{\pm 2.82}$ & $14.27_{\pm 2.81}$ \\
 & DeepSeek & $\underline{66.58_{\pm 2.67}}$ & $50.14_{\pm 0.14}$ & $23.49_{\pm 0.21}$ & $28.44_{\pm 0.19}$ & $30.22_{\pm 0.26}$ & $30.87_{\pm 0.33}$ \\
 & Llama & $81.63_{\pm 0.91}$ & $30.90_{\pm 0.27}$ & $3.42_{\pm 0.13}$ & $6.01_{\pm 0.29}$ & $7.26_{\pm 0.20}$ & $8.61_{\pm 0.47}$ \\ \hline
Eschatocol & Aeneas & $66.18_{\pm 3.36}$ & $44.63_{\pm 0.91}$ & $27.14_{\pm 0.88}$ & $\mathbf{\underline{41.01_{\pm 1.57}}}$ & $\mathbf{\underline{43.42_{\pm 0.56}}}$ & $\mathbf{\underline{47.44_{\pm 1.29}}}$ \\
 & LaTa & $85.57_{\pm 7.68}$ & $38.19_{\pm 1.38}$ & $17.19_{\pm 1.12}$ & $22.01_{\pm 0.81}$ & $23.62_{\pm 1.87}$ & $24.32_{\pm 2.88}$ \\
 & PhilTa & $173.08_{\pm 85.32}$ & $27.93_{\pm 2.48}$ & $9.55_{\pm 2.02}$ & $10.95_{\pm 1.20}$ & $11.46_{\pm 0.93}$ & $11.66_{\pm 1.12}$ \\
 & DeepSeek & $\mathbf{\underline{61.60_{\pm 0.15}}}$ & $\mathbf{\underline{56.07_{\pm 0.12}}}$ & $\mathbf{\underline{31.79_{\pm 0.27}}}$ & $35.41_{\pm 0.31}$ & $37.32_{\pm 0.27}$ & $37.83_{\pm 0.27}$ \\
 & Llama & $80.05_{\pm 1.98}$ & $32.34_{\pm 0.91}$ & $4.63_{\pm 0.51}$ & $7.05_{\pm 0.98}$ & $10.17_{\pm 1.33}$ & $11.78_{\pm 1.27}$ \\ \hline
Rogatio & Aeneas & $84.37_{\pm 3.84}$ & $34.18_{\pm 1.62}$ & $\mathbf{\underline{19.46_{\pm 2.32}}}$ & $\mathbf{\underline{30.45_{\pm 1.23}}}$ & $\mathbf{\underline{32.61_{\pm 0.94}}}$ & $\mathbf{\underline{34.59_{\pm 1.28}}}$ \\
 & LaTa & $91.07_{\pm 7.60}$ & $28.61_{\pm 1.16}$ & $11.71_{\pm 0.79}$ & $16.22_{\pm 1.37}$ & $17.66_{\pm 1.28}$ & $17.84_{\pm 0.94}$ \\
 & PhilTa & $175.18_{\pm 54.96}$ & $17.63_{\pm 2.66}$ & $4.14_{\pm 2.69}$ & $5.59_{\pm 2.00}$ & $5.95_{\pm 2.03}$ & $6.13_{\pm 2.29}$ \\
 & DeepSeek & $\mathbf{\underline{78.62_{\pm 0.98}}}$ & $\mathbf{\underline{36.09_{\pm 0.25}}}$ & $14.59_{\pm 0.50}$ & $17.30_{\pm 0.50}$ & $18.20_{\pm 0.50}$ & $18.92_{\pm 0.79}$ \\
 & Llama & $90.35_{\pm 1.79}$ & $20.56_{\pm 0.61}$ & $1.81_{\pm 0.02}$ & $1.99_{\pm 0.49}$ & $1.99_{\pm 0.49}$ & $2.71_{\pm 0.78}$ \\ \hline
\end{tabular}
}
\caption{Unknown-length setting: results divided by tag (mean $\pm$ 95\% CI, n=5 seeds).}
\label{tab_unknown_divided}
\end{table}

%% file: tab_ablation_llm.tex
\begin{table}[h]
\centering

\label{tab_deepseek_ablation}
\resizebox{0.95\textwidth}{!}{
\begin{tabular}{ll|cccccc}
\hline
Setting & Prompting & CER & Overlap & HR@1 & HR@3 & HR@5 & HR@10 \\ \hline
Known length & Zero-shot & $57.62_{\pm 4.86}$ & $46.10_{\pm 0.06}$ & $20.15_{\pm 0.20}$ & $25.29_{\pm 0.27}$ & $26.20_{\pm 0.24}$ & $26.61_{\pm 0.20}$ \\
 & Few-shot & $49.87_{\pm 0.30}$ & $47.59_{\pm 0.19}$ & $21.75_{\pm 0.33}$ & $25.86_{\pm 0.43}$ & $26.63_{\pm 0.47}$ & $27.24_{\pm 0.60}$ \\
 & Concordance & $56.81_{\pm 5.04}$ & $49.57_{\pm 0.17}$ & $22.92_{\pm 0.28}$ & $28.14_{\pm 0.41}$ & $30.19_{\pm 0.29}$ & $31.34_{\pm 0.16}$ \\
 & Few-shot + concordance & $\mathbf{\underline{47.55_{\pm 0.58}}}$ & $\mathbf{\underline{53.00_{\pm 0.86}}}$ & $\mathbf{\underline{26.38_{\pm 1.12}}}$ & $\mathbf{\underline{31.98_{\pm 1.13}}}$ & $\mathbf{\underline{33.24_{\pm 1.24}}}$ & $\mathbf{\underline{34.28_{\pm 1.71}}}$ \\ \hline
Unknown length & Zero-shot & $100.92_{\pm 6.22}$ & $42.51_{\pm 0.19}$ & $16.75_{\pm 0.30}$ & $20.75_{\pm 0.30}$ & $21.51_{\pm 0.34}$ & $21.92_{\pm 0.30}$ \\
 & Few-shot & $73.09_{\pm 2.63}$ & $44.48_{\pm 0.16}$ & $19.27_{\pm 0.38}$ & $22.49_{\pm 0.38}$ & $23.14_{\pm 0.20}$ & $23.76_{\pm 0.14}$ \\
 & Concordance & $107.37_{\pm 3.21}$ & $46.86_{\pm 0.15}$ & $20.47_{\pm 0.15}$ & $25.41_{\pm 0.11}$ & $26.87_{\pm 0.20}$ & $27.72_{\pm 0.22}$ \\
 & Few-shot + concordance & $\mathbf{\underline{67.15_{\pm 1.59}}}$ & $\mathbf{\underline{49.94_{\pm 0.09}}}$ & $\mathbf{\underline{23.86_{\pm 0.24}}}$ & $\mathbf{\underline{28.41_{\pm 0.22}}}$ & $\mathbf{\underline{30.04_{\pm 0.13}}}$ & $\mathbf{\underline{30.76_{\pm 0.20}}}$ \\ \hline
\end{tabular}
}
\caption{Prompting ablation for DeepSeek on the full test set (mean $\pm$ 95\% CI across 5 rounds).}
\end{table}

%% file: tab_lexical_richness.tex
\begin{table}[H]
\centering
\resizebox{0.4\textwidth}{!}{%
\begin{tabular}{l|ccc}
\hline
Tag & Type-Token Ratio &  Hapax-Token Ratio \\
\hline
Protocol & 0.137 &  6.8\% \\
Eschatocol  & 0.223  & 14.8\% \\
Text Body & 0.129  & 6.7\% \\
Rogationes & 0.310 & 19.7\% \\

\hline
\end{tabular}%
}
\caption{Lexical richness and Hapax Legomena-to-token ratio across notary document sections.}
\label{hapax}
\end{table}

%% file: fig_prompt.tex
\fbox{%
\parbox{0.95\linewidth}{%
You are an expert in medieval Latin paleography and diplomatic documents. You need to restore lacunae in medieval Latin documents (charters, bulls, notarial acts, cartularies, and related diplomatic texts).

The user will provide a passage with a single lacuna marked ***. The string [MISSING] may also appear; treat it as fixed context; do NOT attempt to restore it.

The lacuna may NOT align with word boundaries. *** can begin or end mid-word. Treat the characters immediately before and after *** as literal: restore only what belongs strictly between them.

The expected length of restored text is 10 characters, with letters, punctuations, and spaces included.

Produce the ten most plausible restorations, ordered from most to least confident. The candidates should be genuinely distinct readings, not minor orthographic variants of a single guess.

Output format: a single Python list literal containing exactly ten strings, and nothing else. No code fences, no preamble, no trailing commentary. Each string contains only the characters that replace ***, with no punctuation beyond what the restoration itself requires.

Here are some examples:

Text:

In nomine sancte et individue trinitatis. Ego Iohan***tarius Anno domini

Restored text: ['nes no', 'nis no', 'nes publicus no', 'nes sacri palacii no', 'nes imperiali auctoritate no', 'nes civis no', 'nes dictus no', 'nes dei gratia no', 'nes apostolica auctoritate no', 'nis publici no']

Text:

Anno domini millesimo du***imo sexagesimo tercio die lune

Restored text: ['centes', 'ocentes', 'centess', 'ocentess', 'centis', 'centens', 'ocentis', 'gentes', 'cetes', 'ocentens']

Text:

die lune post fes***ancti Michael[MISSING]

Restored text: ['tum s', 'to s', 'tivitatem s', 'tivitate s', 'tum beati s', 'to beati s', 'tum gloriosi s', 'tum translationis s', 'tum dedicationis s', 'ta s']

Corpus concordances from a reference corpus of similar notarial documents, listed most-common first. Use them as hints; they may be incomplete or not applicable.

The word immediately before *** is "filio". In the corpus it is most often followed by:

  - "quondam"
  
  - "suo"
  
  - "iohannes"

The characters just before *** ("Albert") may be the start of a word. In the corpus, "filio" is most often followed by words beginning with "Albert":

  - "alberto" -> *** would begin with "o"
  
  - "alberti" -> *** would begin with "i"
  
  - "albertus" -> *** would begin with "us"

The word immediately after *** is "et". In the corpus it is most often preceded by:

  - "se"
  
  - "possidendum"
  
  - "sancte"

The characters just after *** ("ti") may be the end of a word. In the corpus, "et" is most often preceded by words ending with "ti":

  - "viginti"  -> *** would end with "vigin"
  
  - "abbati"  -> *** would end with "abba"
  
  - "alberti"  -> *** would end with "alber"

Now restore the following:
...filio Albert***ti et Hermengarde...

Restored text:
}}